\documentclass{ieeetj}

\usepackage{cite}
\usepackage{amsmath,amssymb,amsfonts}
\usepackage{algorithmic}
\usepackage{graphicx,color}
\usepackage{textcomp}
\usepackage{xcolor}
\usepackage[colorlinks=true, citecolor=blue, linkcolor=blue, urlcolor=red]{hyperref}
\usepackage{algorithm}
\def\BibTeX{{\rm B\kern-.05em{\sc i\kern-.025em b}\kern-.08em
    T\kern-.1667em\lower.7ex\hbox{E}\kern-.125emX}}
\AtBeginDocument{\definecolor{tmlcncolor}{cmyk}{0.93,0.59,0.15,0.02}\definecolor{NavyBlue}{RGB}{0,86,125}}

\usepackage{todonotes} 
\usepackage{acronym}
\usepackage{siunitx}
\usepackage[capitalize]{cleveref}
\usepackage{placeins}
\usepackage{booktabs}
\usepackage{overpic}
\usepackage{needspace}
\usepackage{subfigure}
\usepackage{bbm}
\usepackage{xurl}
\usepackage{xspace}

\acrodef{CNN}{Convolutional Neural Network}
\acrodef{CoM}{Center of Mass}
\acrodef{DoF}{Degree of Freedom}
\acrodef{FF}{Feed Forward}
\acrodef{IK}{Inverse Kinematics}
\acrodef{IMU}{Inertial Measurement Unit}
\acrodef{KF}{Kalman Filter}
\acrodef{LUT}{Look-Up Table}
\acrodef{MAE}{Mean Absolute Error}
\acrodef{ML}{Machine Learning}
\acrodef{MLP}{Multilayer Perceptron}
\acrodef{MPC}{Model Predictive Control}
\acrodef{NN}{Neural Network}
\acrodef{PPO}{Proximal Policy Optimization}
\acrodef{PRM}{Probabilistic Roadmaps}
\acrodef{RL}{Reinforcement Learning}
\acrodef{RRT}{Rapidly-exploring Random Trees}
\acrodef{RRT*}{RRT-STAR}
\acrodef{ROS}{Robot Operating System}
\acrodef{SM}{State Machine}
\acrodef{TCN}{Temporal Convolution Network}
\acrodef{THO}{Task Hierarchical Optimization}
\acrodef{CAV}{Center Articulated Vehicle}
\acrodefplural{CAV}[CAVs]{Center Articulated Vehicles}

\newif\ifanonymous
\anonymousfalse 

\DeclareRobustCommand{\anontext}[2]{%
  \ifanonymous
    \textcolor{gray}{[#2]}%
  \else
    #1%
  \fi
}

\def\OJlogo{\vspace{-4pt}$<$Society logo(s) and publication title will appear here.$>$}
\def\seclogo{\vspace{10pt}$<$Society logo(s) and publication title will appear here.$>$}

\makeatletter
\DeclareRobustCommand\onedot{\futurelet\@let@token\@onedot}
\def\@onedot{\ifx\@let@token.\else.\null\fi\xspace}
\def\etal{{et al}\onedot}
\def\etalcite#1{\etal~\cite{#1}}
\makeatother

\def\authorrefmark#1{\ensuremath{^{\textbf{#1}}}}

\renewcommand\thesection{\Roman{section}}

\begin{document}

\markboth{SAHA: Supervised Autonomous HArvester for selective forest thinning}{\anontext{Nan {et al.}}{Anonymous Authors}}

\title{SAHA: Supervised Autonomous HArvester for selective forest thinning}


\author{
\anontext{
    Fang Nan\authorrefmark{1}\textsuperscript{*},
    Meher Malladi\authorrefmark{2}\textsuperscript{*},
    Qingqing Li\authorrefmark{3},
    Fan Yang\authorrefmark{1},
    Joonas Juola\authorrefmark{3},
    Tiziano Guadagnino\authorrefmark{2},
    Jens Behley\authorrefmark{2},
    Cesar Cadena\authorrefmark{1},
    Cyrill Stachniss\authorrefmark{2},
    and Marco Hutter\authorrefmark{1}
}{anonymous authors}

\anontext{
    \affil{Robotic Systems Lab, ETH Zurich, Zurich, Switzerland}
    \affil{Center for Robotics, University of Bonn, Bonn, Germany}
    \affil{Prefor Oy, Helsinki, Finland}
    
    \authornote{* The authors contributed equally to this work. \\
      This project has received funding from the European Union’s Horizon Europe Framework Programme\\under grant agreement No 101070405 (DigiForest).
    }
    \corresp{Corresponding author: Fang Nan (email: {\href{mailto:fannan@ethz.ch}{\textcolor{blue}{fannan@ethz.ch}}})}
}{anonymous affiliations}
}

\begin{abstract}
Forestry plays a vital role in our society, creating significant ecological, economic, and recreational value.
Efficient forest management involves labor-intensive and complex operations.
One essential task for maintaining forest health and productivity is selective thinning, which requires skilled operators to remove specific trees to create optimal growing conditions for the remaining ones.
In this work, we present a solution based on a small-scale robotic harvester (SAHA) designed for executing this task with supervised autonomy.
We build on a 4.5-ton harvester platform and implement key hardware modifications for perception and automatic control.
We implement learning- and model-based approaches for precise control of hydraulic actuators, accurate navigation through cluttered environments, robust state estimation, and reliable semantic estimation of terrain traversability.
Integrating state-of-the-art techniques in perception, planning, and control, our robotic harvester can autonomously navigate forest environments and reach targeted trees for selective thinning.
We present experimental results from extensive field trials over kilometer-long autonomous missions in northern European forests, demonstrating the harvester's ability to operate in real forests. We analyze the performance and provide the lessons learned for advancing robotic forest management.
\end{abstract}


\maketitle

\section{Introduction}

Forests cover roughly one-third of the Earth's land area~($\approx$\,4.06\,billion hectares), providing critical habitat for biodiversity, supporting rural livelihoods, and sustaining a substantial bioeconomy~\cite{feng2016rse}.
Beyond timber production, vegetated ecosystems play a major role in climate regulation by absorbing on the order of one-third of annual anthropogenic CO$_2$ emissions~\cite{dye2024cbm}.
Such essential forest functions translate into enormous demand for forest operation and management~\cite{dannunzio2015fem,giffen2025fem}.
As scientific and public understanding of forest ecosystems has deepened, requirements on forest management have broadened: timber production must now be balanced with biodiversity conservation, carbon storage, recreation, and other societal values~\cite{Marchi18SustainableForest}.
Meeting these competing goals requires more selective, precise operations and more detailed planning, which in turn increases labor requirements and the need for specialized skills.

At the same time, field execution of forestry tasks remains constrained by safety and workforce challenges.
Logging is among the most hazardous occupations; in the United States the fatal work-injury rate for logging workers reached 98.9 per 100,000 full-time equivalent workers in 2023, compared with 3.5 per 100,000 for all industries~\cite{U.S.BureauOfLaborStatistics24CivilianOccupations}.
Many regions also face a declining and aging forestry workforce—for example, Europe experienced roughly an 18\% decline in forestry employment between 2008 and 2016, with persistent labor shortages noted in more recent reports~\cite{sporcic2025forests}.

\begin{figure*}[t]
  \centering
  \includegraphics[trim=2cm 0 0 0, clip, width=\linewidth]{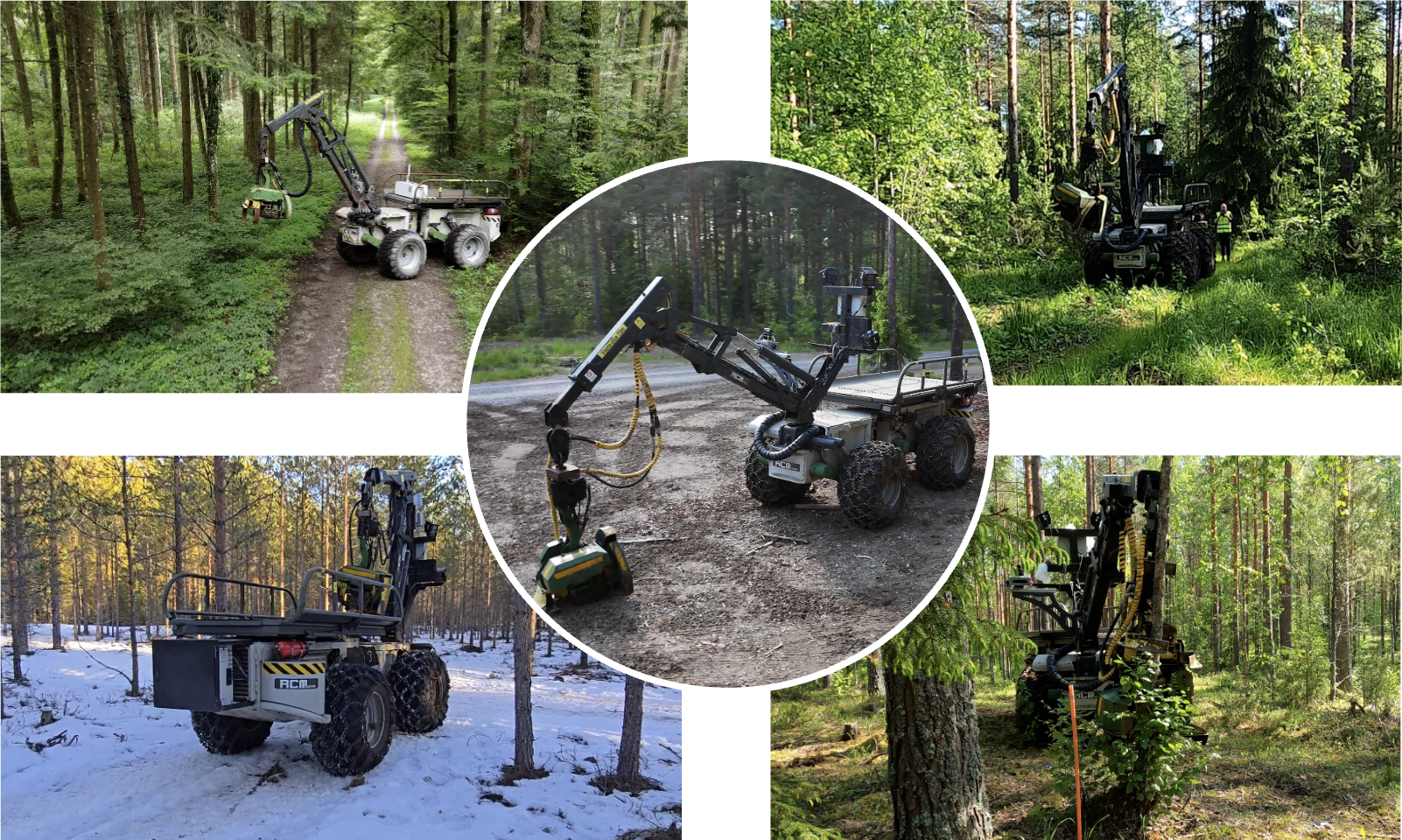}
  \caption{The developed autonomous harvester system has been deployed in real-world experiments across various terrains and seasons. It can navigate along service trails (top left) and traverse cluttered forests (top right) during different seasons (bottom left), and reach and grasp selected trees for thinning (bottom right).}
  \label{fig:site-map}
\end{figure*}

Consequently, automation technologies for forestry are in high demand, not only because they can remove people from the most dangerous tasks and thus make forestry jobs more attractive, but also because they offer consistent, repeatable execution that enables more precise operations.
In particular, autonomous robots for forest applications have received increasing attention from both academia and industry.
While using robots to map and survey forests has been successfully demonstrated~\cite{Freissmuth24OnlineTree,Mattamala25BuildingForest}, carrying out fully automated cutting in forests is uniquely difficult.
Perception is challenging due to dynamic occlusions (e.g., foliage, branches), repetitively textured scenes, and limited visibility; although mobile laser scanning using Simultaneous Localization and Mapping~(SLAM) has advanced~\cite{malladi2025icra}, inventory-grade mapping still contends with occlusions and drift in complex stands~\cite{Wu25ApplicationSLAMbased}.
Under-canopy GNSS is often degraded by multipath effects and occlusion; even modern low-cost, dual-frequency RTK systems suffer accuracy losses beneath dense canopy~\cite{Cateanu24PerformanceEvaluation}.
Mobility and manipulation amplify these issues: heavy hydraulic machines need to operate on deformable, sloped terrain with varying traction and contact-rich interactions, requiring closed-loop perception and control under uncertainty.

Recent systems~\cite{Zhang21AutonomousExcavator, Jelavic22AutonomousRobotic, LaHera24ExploringFeasibility} demonstrate that field-deployable autonomy is within reach but not yet general across sites, seasons, and tasks.
Autonomous excavators now perform material handling tasks in realistic settings~\cite{Zhang21AutonomousExcavator}, and initial attempts at unmanned forestry operations have been reported for harvesting~\cite{Jelavic22AutonomousRobotic} and forwarding~\cite{LaHera24ExploringFeasibility}.
However, demonstrations so far have often been limited to structured or pre-surveyed areas without complex terrain, unknown obstacles, heavy undergrowth, or dense tree cover.
These simplifications eliminate many challenges associated with perception and navigation in cluttered environments often encountered in forestry, particularly in precision thinning operations where the forest is planted densely.

In this paper, we present a system that targets specifically the forest thinning task in real-world forest environments.
Specifically, the contributions of this work are:
\vspace{-2mm}
\begin{itemize}
  \item The development of SAHA (Supervised Autonomous HArvester), a robotic small-scale autonomous harvester designed for selective forest thinning. This system is a modified 4.5-ton harvester platform with key hardware changes for perception and automatic control.
  \item The realization of learning- and model-based approaches for precise control of the hydraulically actuated arm and active chassis for accurate gripper placement and stable driving through complex terrain.
  \item The development of a robust navigation system that enables the harvester to negotiate cluttered environments and reach selected trees for thinning. This includes reliable odometry and traversability estimation modules to support autonomous navigation in forests.
  \item The integration and field tests of the complete system in real forests in northern Europe, demonstrating autonomous operation in realistic forestry environments.
\end{itemize}

\section{Related Work}

\subsection{Robotics in Forestry}
Robotics and automation in forestry continues to attract significant attention, driven by the need to enhance safety, efficiency, and sustainability in forest operations.
Various robotic systems have been developed for tasks such as forest mapping, tree monitoring, log extraction, and autonomous harvesting.

Many robotic platforms have been proposed for forest mapping and inventory applications~\cite{Pierzchala18MappingForests,Tremblay20AutomaticThreedimensional,Freissmuth24OnlineTree,Mattamala25BuildingForest}.
These systems often utilize LiDAR and camera data to create detailed 3D maps of forest environments, from which individual trees can be classified and measured.
Many of these systems are ground-based, using wheeled locomotion to navigate through forest terrain~\cite{Pierzchala18MappingForests,Tremblay20AutomaticThreedimensional}.
With recent advancements in legged locomotion, particularly over challenging terrain~\cite{Miki22LearningRobust, Frey23FastTraversability}, there is growing interest in the use of legged robots for forestry applications due to their ability to navigate uneven and cluttered terrains~\cite{Mattamala25BuildingForest}.
Recent works also consider combining ground-based and aerial data to improve mapping quality, complementing the under-canopy perspective of ground robots with above-canopy information~\cite{Casseau24MarkerlessAerialTerrestrial}.


Although robotic forest surveying has seen many successful examples~\cite{Murtiyoso24VirtualForests, malladi2024icra, malladi2025icra}, robotic execution of forestry operations remains uniquely difficult.
The first unmanned full-size robotic system for autonomous forestry operations is reported by Jelavic~\etalcite{Jelavic22AutonomousRobotic}, who modify a 12-ton hydraulic excavator~\cite{Jud21HEAPAutonomous} to demonstrate autonomous precision cutting operations in a forest.
This system is equipped with appropriate sensors to scan a forest and build a point cloud map, from which individual trees are segmented geometrically.
Once a tree is selected, the system autonomously navigates to the tree and positions its cutting head.
The work by Jelavic~\etalcite{Jelavic22AutonomousRobotic} is the first demonstration of a full-size autonomous forestry machine in literature, but the system is limited to operating in a small, pre-surveyed area without complex dynamic obstacles.
The experiment takes place in a sparse forest without undergrowth and obstacles and only considered trees near the service road.
These simplifications eliminate many challenges associated with perception and navigation in cluttered environments often encountered in thinning operations.

In a related direction, La Hera~\etalcite{LaHera24ExploringFeasibility} present an autonomous forwarder capable of autonomous log transport in a forest environment.
In this work, an unmanned forwarder navigates predefined paths along forest roads and uses visual detection to locate logs left on the roadside by a previous harvesting operation.
When a log is detected, the forwarder stops and uses its crane to pick up the log and load it onto its bunk.
While La Hera~et~al.~test the system in a real forest operation site, the navigation remains limited to driving along forest roads, where GNSS localization is available and obstacle density is low.

More generally, automation solutions for heavy machinery are being actively researched in the context of construction and logistics applications.
Successful field deployments of autonomous construction machinery have been reported for excavation~\cite{Terenzi24AutonomousExcavation,Egli24ReinforcementLearningBased}, material handling~\cite{Spinelli25LargeScale}, and construction~\cite{Johns23FrameworkRobotic}.
Forestry applications, however, present additional challenges in perception and navigation due to the unstructured, cluttered environments.

\subsection{Perception in Forest Environments}

Perception in forestry contexts has used LiDAR-based approaches for forest mapping and inventory~\cite{wulder2012rse}, as well as camera-based approaches for visual scene understanding~\cite{schiefer2020jprs,fortin2022iros, Frey23FastTraversability}.
Data collected by robotic surveying platforms has proven useful for various forestry applications~\cite{Murtiyoso24VirtualForests}.
Early studies use terrestrial laser scanning~\cite{du2019rs,hannah2022essd,burt2019mee} and airborne LiDARs~\cite{dalponte2016mee,sun2022fps,roussel2020rse,li2012pers} for forest surveys, enabling estimation of metrics such as tree diameter at breast height~\cite{liang2016jprs}, canopy height~\cite{liang2018jprs}, and biomass~\cite{gonzalez2018mee}.
More recently, mobile laser scanning platforms such as handheld devices~\cite{donager2021rs}, backpack-mounted systems~\cite{malladi2024icra}, UAVs~\cite{puliti2023arxiv}, and quadrupeds~\cite{Mattamala25BuildingForest} have demonstrated viability for forest mapping.
These platforms can support offline, high-fidelity reconstruction for detailed analysis~\cite{malladi2025icra}, as well as online mapping for estimating tree inventory during field operations~\cite{Freissmuth24OnlineTree}.

Initial methods for tree detection and segmentation primarily relied on geometric techniques such as clustering point clouds~\cite{donager2021rs} or analyzing rasterized canopy height models~\cite{dalponte2016mee}.
Deep learning has enabled more robust semantic and panoptic segmentation approaches~\cite{krisanski2021rs}, though these methods depend heavily on annotated datasets for training.
Several forestry-specific datasets have been developed recently~\cite{malladi2025icra, puliti2023arxiv, hannah2022essd, vidanapathirana2024ijrr}.
However, compared to other domains~\cite{cordts2016cvpr, behley2021icra}, these datasets remain limited in terms of sensor variety, environmental diversity, scale, and annotation complexity.

State estimation in unstructured forest environments is challenging, particularly in GNSS-denied or cluttered settings.
Recent forestry datasets employ robust reference trajectory pipelines to achieve high-quality mapping results.
The DigiForests dataset~\cite{malladi2025icra} provides extensive semantic annotations of forest scenes along with spatially aligned trajectories and mapping results from data collected across multiple forest sites over three seasons.
Malladi~\etalcite{malladi2025icra} obtain the trajectories through an offline pose-graph optimization, leveraging the VILENS system~\cite{wisth2023tro} and loop closures detected between recording sessions using learning-based place recognition~\cite{oh2024iros}.
Such loop closure detection in forest SLAM is an active area of research, with learning-based approaches showing promising results~\cite{shen2025arxiv, oh2024iros}.
Similarly, the WildScenes~\cite{vidanapathirana2024ijrr} dataset utilizes a combination of the Wildcat~\cite{ramezani2022arxiv} continuous-time LiDAR-inertial SLAM system and offline GNSS-integrated bundle adjustment to provide its reference trajectories.
The TreeScope~\cite{cheng2023arxiv} dataset provides pose trajectories for its data obtained using Faster-LIO~\cite{bai2022ral}.
Both datasets also provide semantic annotations for trees.
Such datasets show potential for advancing both semantic scene interpretation and state estimation methods in challenging forest environments.

Mobile robots in forest environments also require an understanding of traversability,~i.e.,~they should be able to perceive which parts of the forest are accessible and navigable.
Geometric approaches to estimate traversability typically analyze 3D data such as point clouds or elevation maps, where terrain slope, roughness, or obstacle density are used to infer navigability~\cite{cao2022icra, dixit2024fr}.
While effective for structured environments, these methods often fail in forests due to vegetation and occluded ground surfaces that distort geometric cues~\cite{Frey23FastTraversability}.
Learning-based segmentation approaches extend this by predicting traversable regions based on semantics or appearance~\cite{yang2021icra}, enabling navigation in unstructured off-road settings.
However, such methods rely on annotations, which are scarce and environment-specific, limiting their generalization capabilities~\cite{bradley2015ricirs, shaban2022rl}.
To overcome these limitations, recent self-supervised learning frameworks infer traversability directly from motion cues using visual and proprioceptive feedback, enabling online adaptation in natural environments~\cite{Frey23FastTraversability, mattamala2025ar}.

\subsection{Planning in Forest Environments}
Autonomous planning in forests and similarly unstructured environments faces multiple challenges, including dense and irregular obstacles, limited or no prior maps, and platform-specific kinematic constraints,~e.g.,~\ac{CAV} platforms commonly used in working machines.
These factors favor agile local planners driven by onboard perception and frequent replanning rather than global, map-dependent methods.

Sampling-based and grid-search methods perform effectively in structured environemnts and urban settings~\cite{Gammell14InformedRRT,Hart68FormalBasis,HwanJeon13OptimalMotion}.
However, their reliance on consistent global maps limits robustness in unknown, dynamic forest scenes.
Optimization-based approaches~\cite{Jelavic21CombinedSampling,Guo08OptimalTrajectory} can adapt online to dynamics but become computationally intensive when dealing with complex vehicle models and cluttered scenes, hindering real-time use in off-road applications.
Learning-based methods that directly map sensory input to actions can offer high reactivity. Yet, encoding hard kinematic and feasibility constraints can be difficult, risking trajectories that violate nonholonomic limits in tight clutter~\cite{Chen23EndtoendAutonomous}.
This weakness is problematic for heavy off-road vehicles maneuvering among trees, brush, and uneven terrain.

Precomputed primitives enable fast, kinodynamically feasible planning, making them well-suited to the unknown, obstacle-rich settings typical for forests.
Successes span exploration and navigation tasks for autonomous aerial and ground vehicles~\cite{Zhang20FalcoFast,Dharmadhikari20MotionPrimitivesbased,Jarin-Lipschitz21DispersionminimizingMotion,Low21PROMPTProbabilistic}.
Primitive design options include atomic~\cite{Dubins57CurvesMinimal,Reeds90OptimalPaths}, state-lattice and control-sampling schemes~\cite{Pivtoraiko11KinodynamicMotion}, and data-driven variants~\cite{Deng18LearningbasedHierarchical}.
Receding-horizon planners leveraging such primitives have shown agile navigation in obstructed environments~\cite{Zhang20FalcoFast}, aligning with the demands of forest settings.

A notable gap is the usage of specialized primitives for \ac{CAV}s that respect articulation limits and minimum turning radii.
These are crucial for negotiating narrow gaps between trees encountered in forest deployments.
While \ac{CAV} kinematics and dynamics have been studied~\cite{DeSantis97ModelingPathtracking,Corke01SteeringKinematics}, existing works on planning and control of \ac{CAV}s use simplified models and methods adopted for car- or bicycle-like vehicles, which limits their performance in dense, unstructured environments~\cite{Nayl13ModelingControl}.


\subsection{Control of Heavy Machinery}
The control of heavy machinery, particularly systems with hydraulic actuation, presents unique challenges due to the nonlinear, time-varying dynamics and the presence of significant delays and uncertainties.
As a result, early works in model-based control for hydraulic machines often relied on simplified models and could not achieve high accuracy~\cite{Ha02RoboticExcavation}.
Recent hardware advances partially address these challenges through the integration of high-quality sensors and valves.
For example, Hutter~\etalcite{Hutter15OptimalForce,Hutter17ForceControl} integrate custom servo-valves into a hydraulic legged excavator to achieve precise force control for chassis balancing.
Such actuators are also used later on the excavator's arm for force-controlled grading and excavation tasks~\cite{Jud17PlanningControl}.
Compared to standard proportional valves typically used in heavy machinery, these actuators offer improved control accuracy and bandwidth, although their high manufacturing and maintenance costs limit practical applications.

More recently, data-driven control methods have emerged as a promising alternative and have received increasing attention.
A notable example is presented by Egli~\etalcite{Egli22GeneralApproach}, who train a deep neural network to model the hydraulic dynamics of an excavator arm and use it for reinforcement learning-based training of an accurate controller.
Along similar lines, Lee~\etalcite{Lee22PrecisionMotion} and Weigand~\etalcite{Weigand21HybridDataDriven} use different model architectures and report better sample efficiencies.
These methods, however, focus on controlling a single machine that the model is trained for.
Nan~\etalcite{Nan24LearningAdaptive} explore the generalization of learned models and controllers across different machines, where they use a latent-space adaptation method to deploy a single controller across multiple hydraulic machines through online adaptation and achieve comparable performance to machine-specific controllers.
Latest research has also shown that it is possible to learn control policies directly on the real machine using efficient online learning methods~\cite{nan2025efficient}.

\section{Method}
\label{sec:method}

\begin{figure*}[!t]
  \centering
  \includegraphics[width=0.9\textwidth]{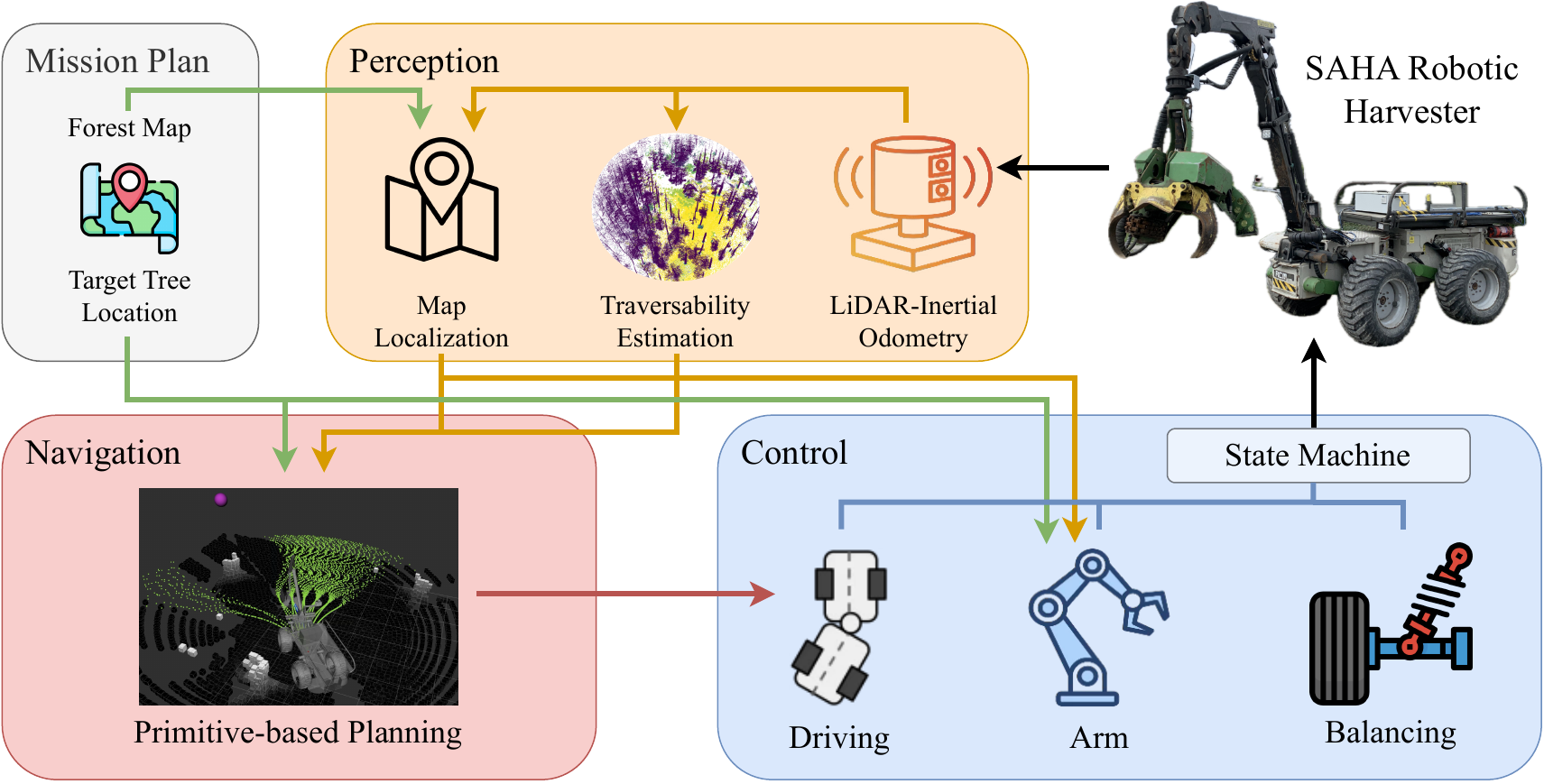}
  \caption{Overview of the SAHA system. It localizes itself initially in a prior map and uses onboard odometry for position tracking. A traversability estimation module identifies navigable terrain, guiding path planning by the motion primitive-based planner responsible for collision avoidance during navigation. Driving, arm control, and chassis balancing controllers manage autonomous movement and stability. A state machine coordinates transitions between navigation and cutting operations.}
  \label{fig:saha_system_overview}
\end{figure*}

\subsection{System Overview}

We develop the SAHA robot for the first thinning task, which refers to the initial removal of selected smaller or suppressed trees in a young stand to provide remaining trees with more space, light, and resources for optimal growth.
This task requires highly selective cutting to remove specific trees while maintaining overall forest health.
This process is both labor-intensive and critical for enabling natural regeneration.
To automate the first thinning task, SAHA features a compact hardware design that minimizes soil compaction and enhances maneuverability within dense forests.
SAHA's software stack enables robust state estimation, reliable navigation in cluttered forest environments, and precise control during cutting operations.

In~\cref{fig:saha_system_overview}, we illustrate the overall system architecture of the SAHA robot.
SAHA is built upon a small harvester platform with active chassis capabilities, and we develop a robust perception, navigation, and control pipeline to enable autonomous operation in challenging forest environments.
We assume that a forest inventory is available, potentially acquired through forest mapping conducted by aerial and ground-based robots~\cite{Mattamala25BuildingForest}.
Target trees for thinning are assumed to be provided as well, either manually by an operator or through a decision support system~\cite{malladi2025icra}.

Upon receiving a mission plan consisting of a global map of the forest and the location of the next target tree, SAHA first localizes itself in the global map.
It then navigates autonomously toward the selected tree, utilizing a motion primitive-based planner while avoiding any obstacles.
As the thinning operation involves driving in unstructured dense forests without clear distinction between drivable and non-drivable areas, we do not rely on global path planning but rather use a local planning approach that is designed for unknown environments.
During autonomous navigation, the system employs a LiDAR-inertial odometry to maintain reliable state estimation, and a learning-based traversability classifier analyzes the LiDAR data to identify navigable terrain.
Upon reaching the target tree, SAHA stops and extends its hydraulic arm to grasp the tree based on the given position of the tree in the global map.
Throughout the entire operation, driving, arm, and chassis balancing controllers ensure precise path following, accurate arm motion, and stable chassis pose.
A state machine negotiates the switch between autonomous navigation and cutting operation, and executes the corresponding controller commands.
Once the tree is securely held in the harvesting head, for safety assurance, a human supervisor remotely triggers the cutting operation.
Throughout the integrated experiments reported in this work, a human supervisor monitored the system within line of sight and could intervene when necessary, by overriding the autonomous commands with a joystick controller.

\subsection{Hardware Platform and Sensor Integration}
\label{ssec:method_hardware}

\begin{figure}[t]
  \centering
  \includegraphics[width=0.85\linewidth]{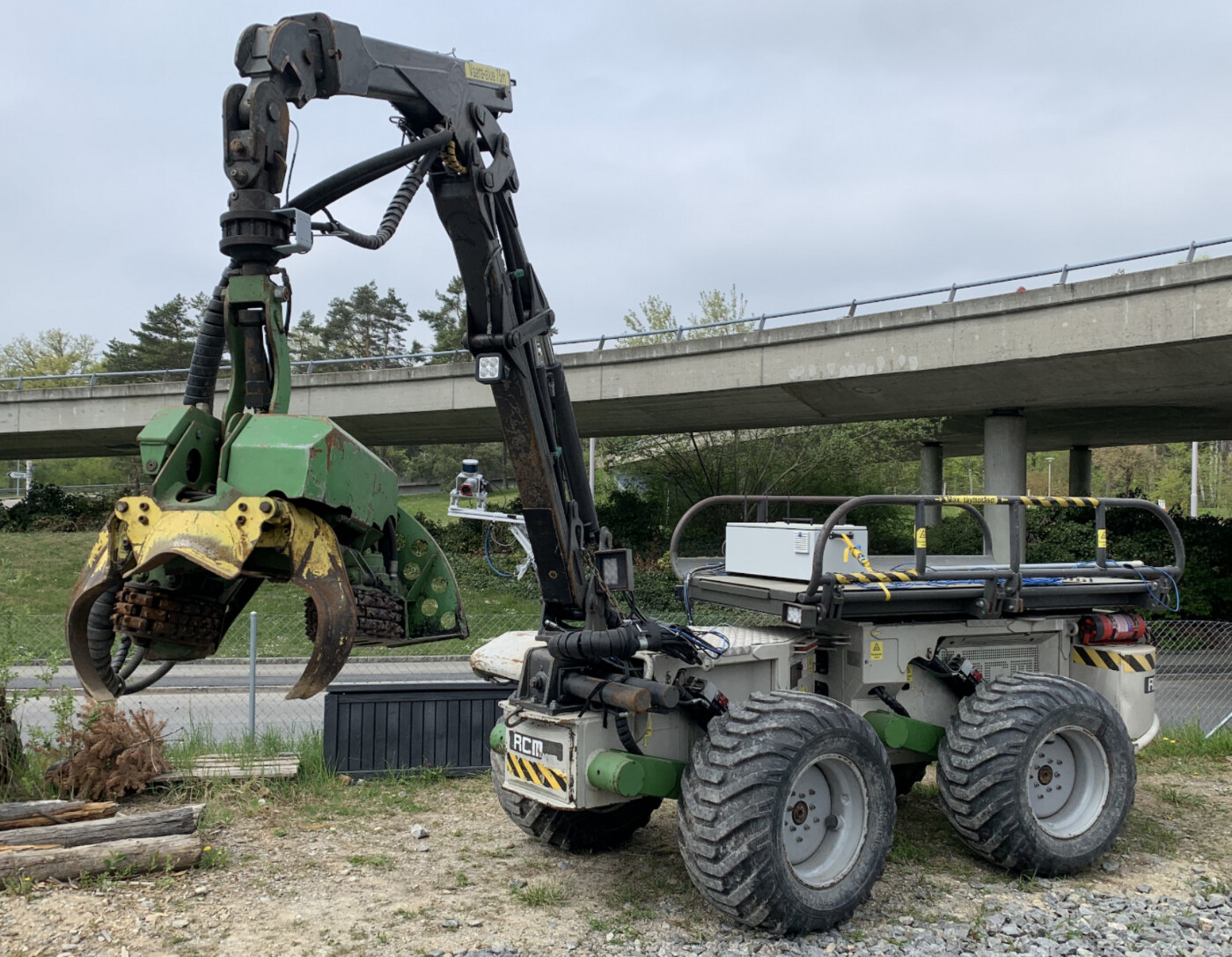}
  \caption{The depicted SAHA robot is a versatile and autonomous forest logging machine. Its lightweight platform compared to other harvesting machines simplifies forest logistics and helps reduce environmental impact through decreased soil compaction.}
  \label{fig:saha_hw_picture}
\end{figure}

The SAHA robot, as shown in~\cref{fig:saha_hw_picture}, is an autonomous forest logging machine designed based on Harveri~\cite{Jelavic22HarveriSmall}, a 4.5-ton machine that combines both cutting and forwarding capabilities within a single, lightweight platform.
This integrated approach not only simplifies the logistics of forest operations but also reduces soil compaction and overall environmental impact.
Unlike heavier machines that require additional support equipment, SAHA's design minimizes soil damage, which is a particularly valuable feature for first thinning operations in forests.

\begin{figure}
  \centering
  \includegraphics[trim={5cm 8cm 0 5cm},clip,width=0.85\linewidth]{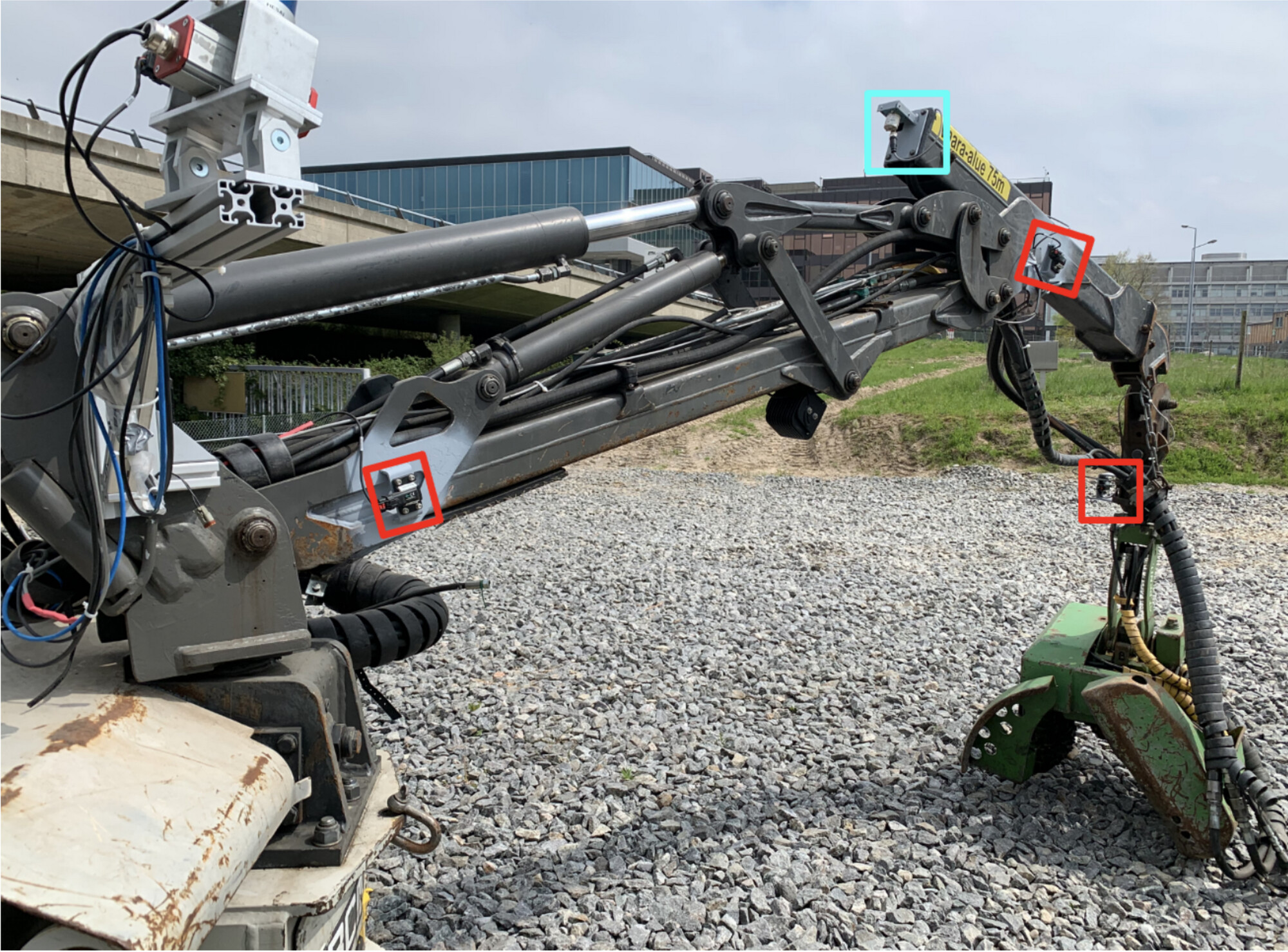}
  \caption{Multiple proprioceptive sensors are installed on the SAHA arm. The red boxes mark the position of IMU sensors, and the cyan box marks the wire draw encoder.}
  \label{fig:saha_hw_proprioceptive_arm}
\end{figure}

\begin{figure}[t]
  \centering
  \includegraphics[trim={0 0cm 8cm 5cm},clip,width=0.85\linewidth]{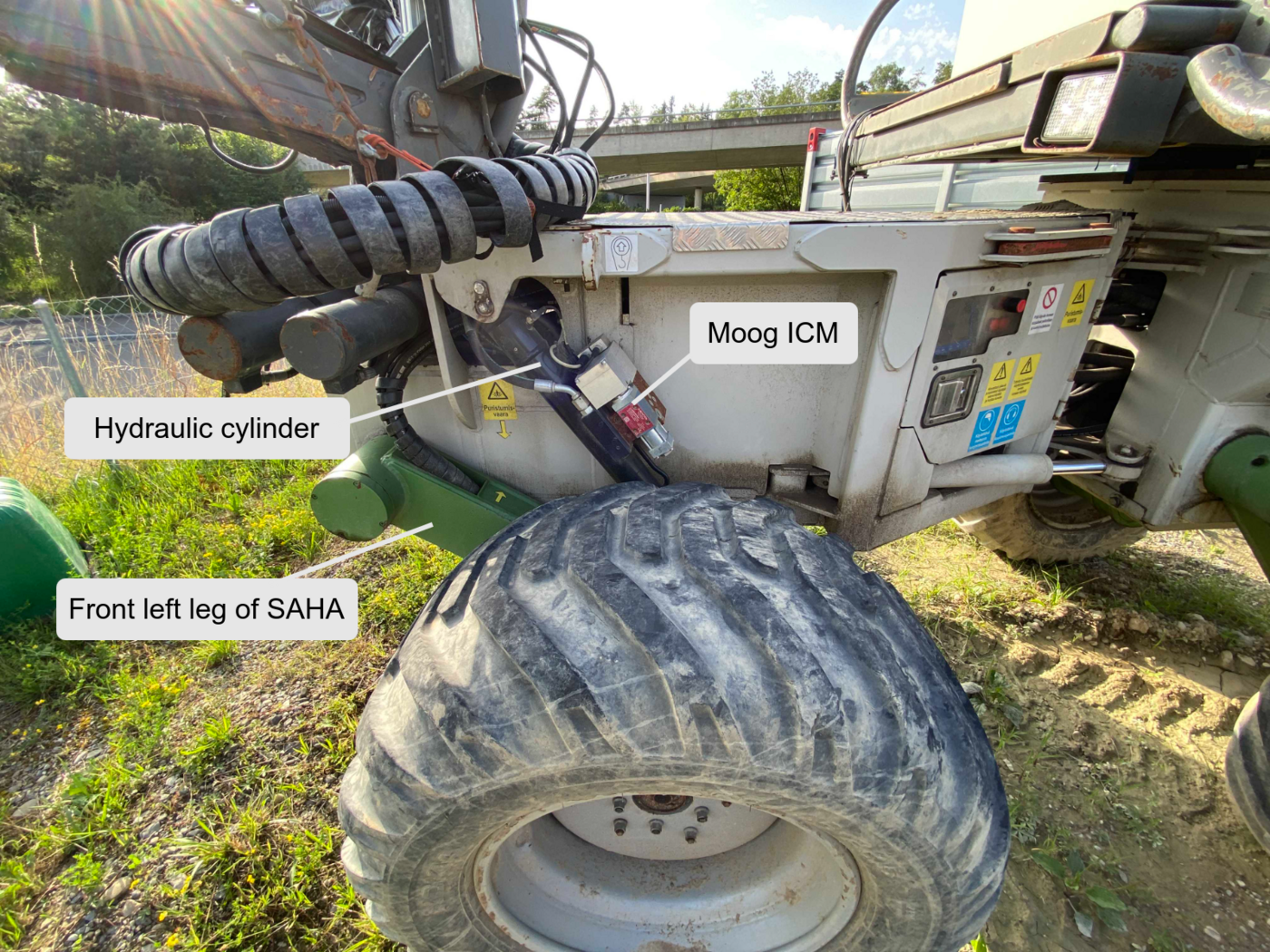}
  \caption{The integrated control module mounted on the left front support cylinder of SAHA.}
  \label{fig:saha_hw_moog_icm}
\end{figure}

Various proprioceptive sensors including Inertial Measurement Units (IMU) and joint encoders are integrated into SAHA, for state estimation of the chassis and the arm.
\cref{fig:saha_hw_proprioceptive_arm} depicts the sensors installed on the arm of the SAHA.
We install the ACEINNA MTLT 335D IMUs on each segment of the arm to provide position and velocity measurements of the rotating joints.
The telescopic boom is equipped with a SICK BCG05 wire draw encoder to measure the extension of the boom.
An SBG Ellipse-A IMU is installed on the chassis to measure the pitch and roll of the machine, and a Gefran GRN hall sensor is used to measure the steering angle between the front and back parts of SAHA.
The active chassis joints are equipped with K\"ubler Sendix M3678A rotary encoders and Wika D20 pressure sensors to provide position and force feedback for active chassis control.
We evaluate two different approaches for modifying the chassis for balancing control (see Sec.~\ref{ssec:method_control}).
The first uses an integrated control module from Moog on each leg joint, similar to the one developed by Hutter~\etalcite{Hutter17ForceControl}, that integrates an internal servo valve for high-bandwidth closed-loop force control.
The second approach uses standard rotary encoders on the chassis joints, along with external pressure sensors on the hydraulic cylinders to provide force feedback and standard proportional valve control.
The installation of the integrated control module can be seen in~\cref{fig:saha_hw_moog_icm}.

\begin{figure}[t]
  \centering
  \subfigure[Front side]{
    \centering
    \includegraphics[width=0.4\linewidth]{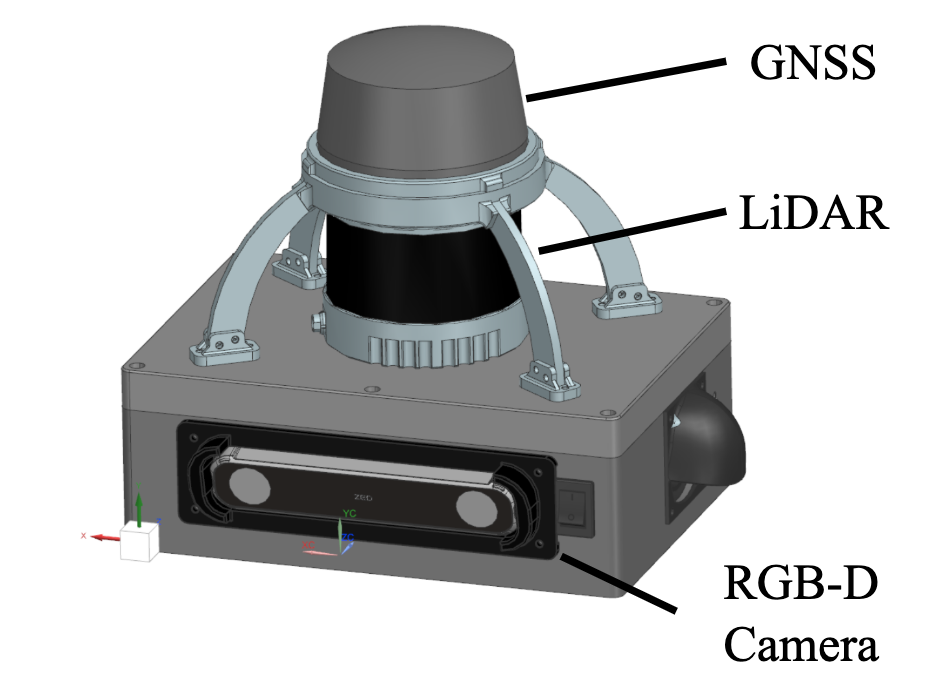}
  }
  \subfigure[Back side]{
    \centering
    \includegraphics[width=0.5\linewidth]{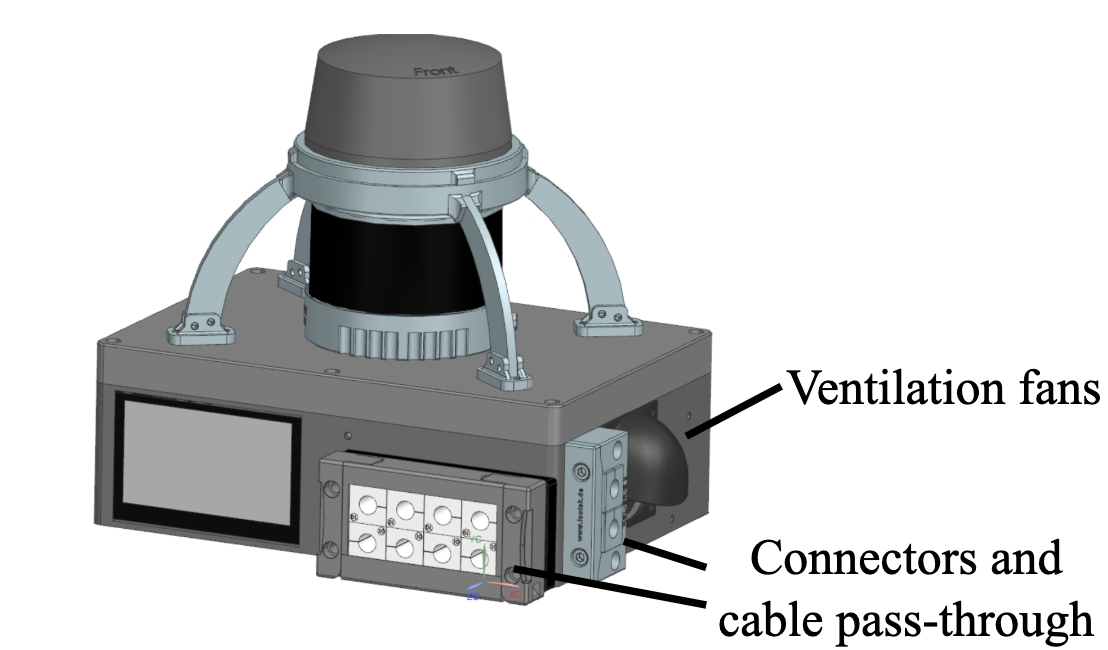}
  }
  \subfigure[Perception kit mounted on SAHA]{
    \centering
    \includegraphics[width=0.6\linewidth]{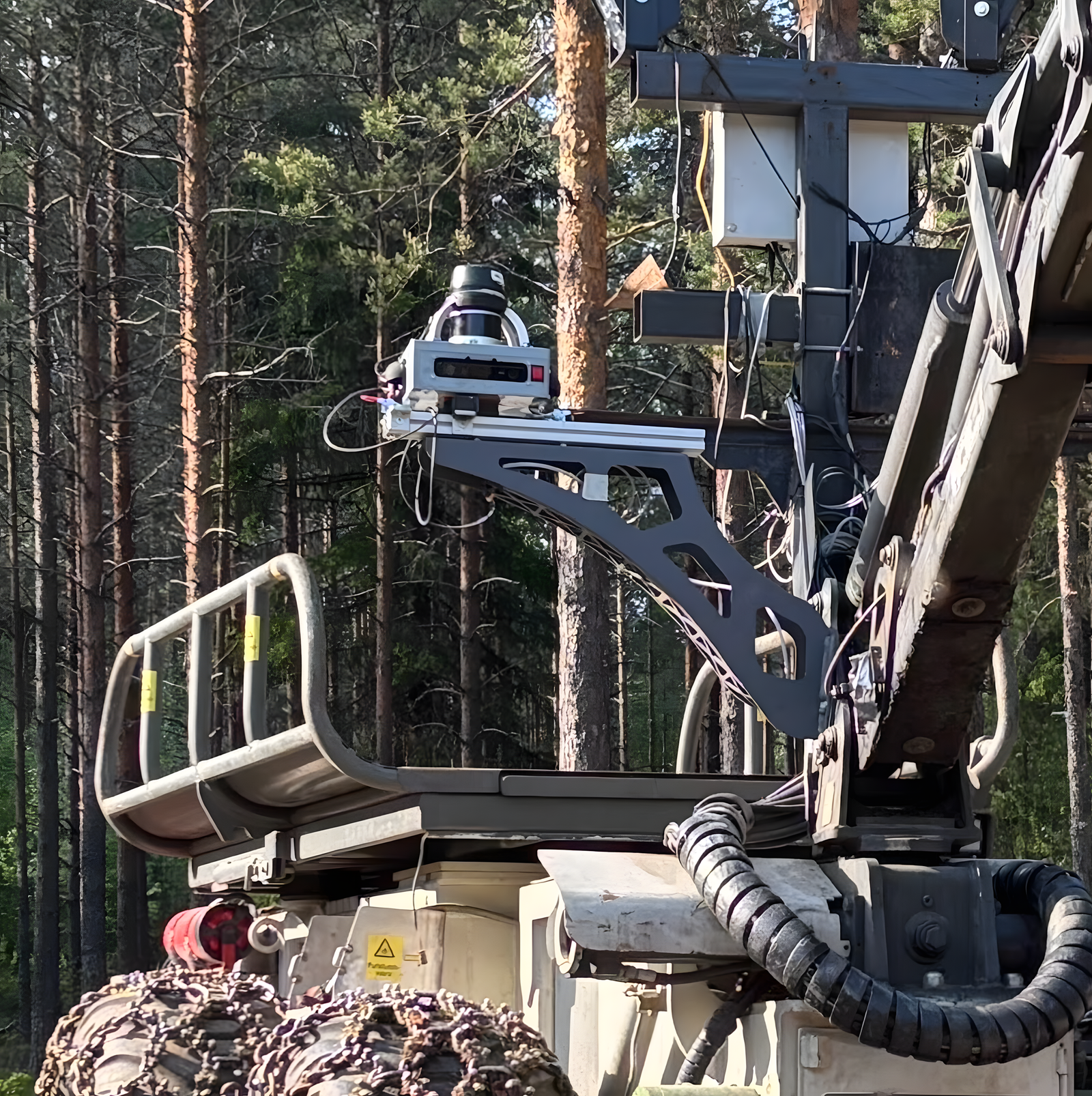}
  }
  \caption{A flexible perception payload is used on the SAHA which integrates a LiDAR, RGB-D camera, IMU and GNSS. (a) and (b) depict the design of the sensor payload, and (c) shows how the payload is mounted on SAHA.}
  \label{fig:saha_hw_hpk}
\end{figure}

\begin{figure*}[t]
  \centering
  \subfigure[Onboard PC on SAHA]{
    \centering
    \includegraphics[width=0.3\linewidth]{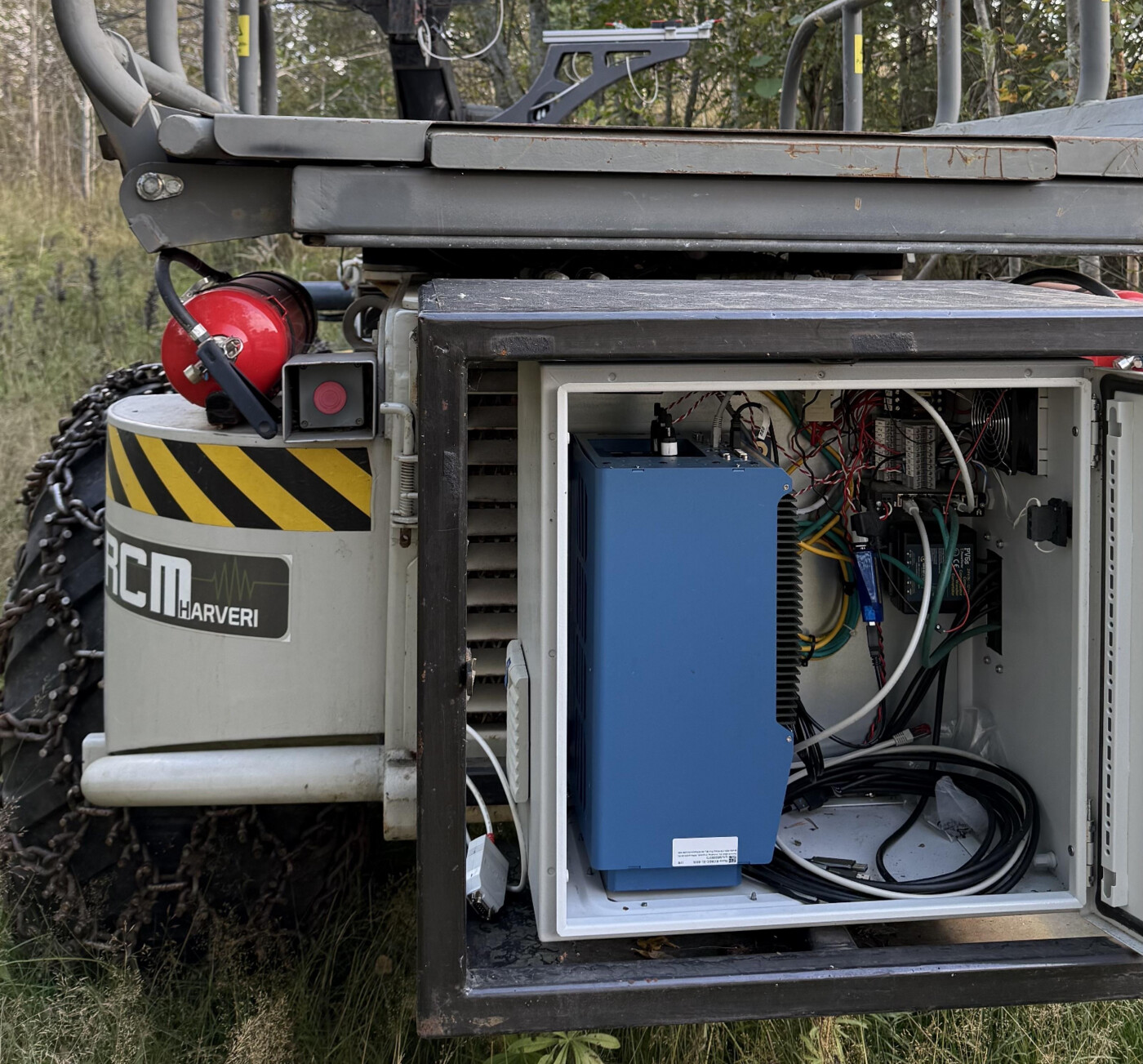}
    \label{fig:saha_hw_onboard_pc}
  }
  \subfigure[Hardware communication diagram of SAHA]{
    \centering
    \includegraphics[width=0.6\linewidth]{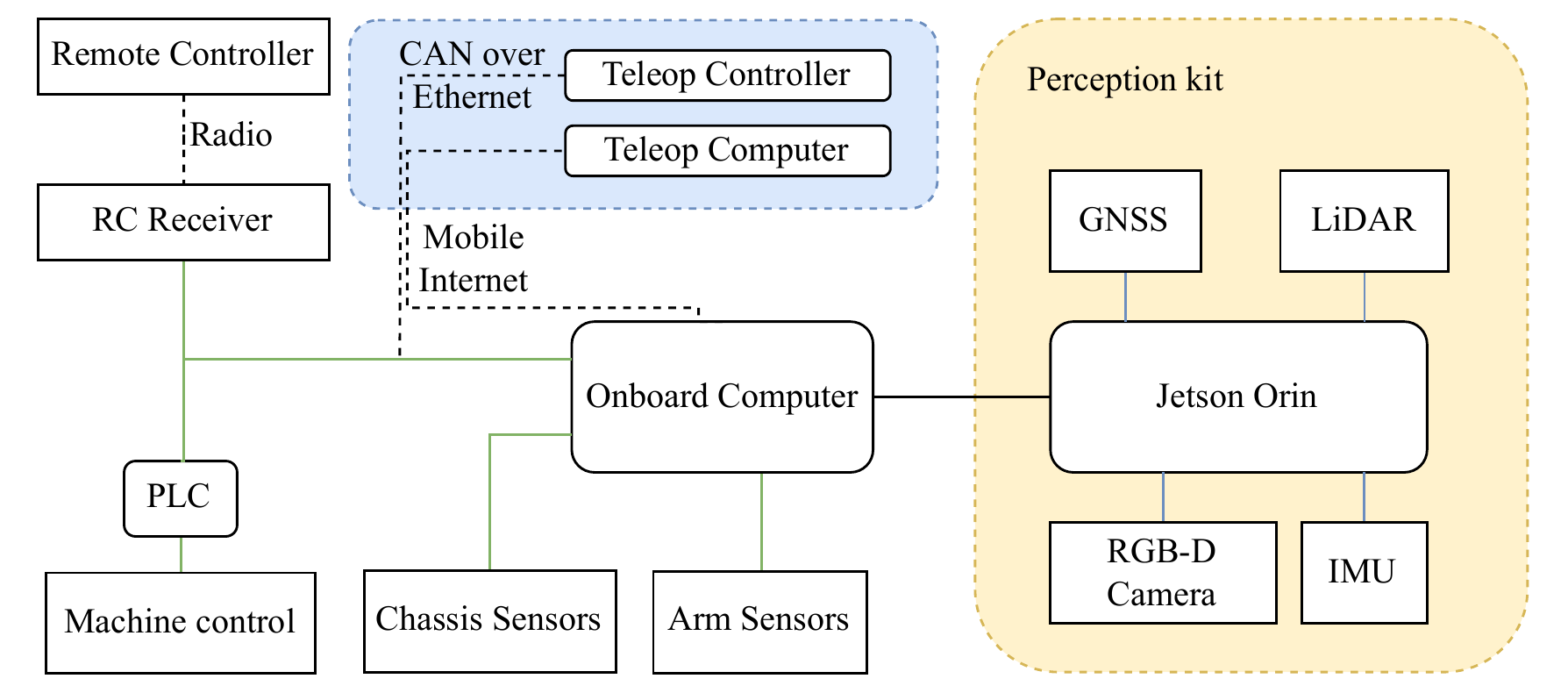}
    \label{fig:saha_hw_communication_diagram}
  }
  \caption{Onboard PC and hardware communication diagram of SAHA. (a) The onboard PC installed in a rugged case in the back of SAHA. (b) A diagram outlining the electronics on the SAHA robot and their connections. In the figure, each black line represents an Ethernet connection, each green line represents a CAN bus connection, and each dashed line represents a wireless connection. The blue box represents the teleoperation station, and the yellow box represents the perception kit.}
\end{figure*}

For environmental perception, we developed a flexible perception payload, as shown in~\cref{fig:saha_hw_hpk}.
The payload integrates several sensors: a top-mounted Hesai XT32 3D LiDAR, a front-facing Stereolabs ZED 2i RGB-D camera tilted 13° down, an Xsens MTI-100 IMU, and a GNSS module. All sensors are time-aligned and connected to an Nvidia Jetson Orin AGX housed in a rugged ABS enclosure.
Components are fixed inside the enclosure with medium-density fiberboard and 3D-printed mounts, with laser-cut apertures providing precise openings for cooling and cable routing.
A dedicated PCB distributes power to all devices, supplied either from the platform's 24\,V rail or an external battery.
The estimated power consumption is 52\,W, for which thermal analysis indicated that passive dissipation alone is inadequate due to the dense component layout.
Therefore, we installed dual 60\,mm fans to maintain the internal temperature well below a 45°\,C limit during sun-exposed operation.

The sensor kit is designed to withstand outdoor working conditions in forests.
Impact robustness of the LiDAR is ensured by a shield composed of an SLS-printed PA-12 top shroud and diagonal aluminum struts.
An externally mounted camera bracket prevents push-in failures while preserving the camera's field of view. The GNSS module is on a breakaway mount at the top.
Fan inlet and outlet covers are shaped to prevent water ingress from above.
All cable feed-throughs use rubber grommets for ingress protection, making the enclosure resistant to rain and dust.

The perception kit is connected to an onboard PC via a single Ethernet cable. This onboard PC, a Neousys Nuvo-8108GC industrial computer with GPU acceleration, also communicates with the proprioceptive sensors, chassis valves, and electric proportional valves actuating the SAHA arm.
The PC is installed at the tail of SAHA in a protected cage, as shown in~\cref{fig:saha_hw_onboard_pc}.
It handles all the computation for terrain analysis, navigation, and control. The odometry subsystem runs on the Jetson in the perception payload.
As illustrated in~\cref{fig:saha_hw_communication_diagram}, the PC interfaces with the perception kit and other components via a series of communication protocols.
The harverster originally used a radio-based remote control mechanism, and this is retained as a redundant backup.
Additionally, the onboard PC enables remote teleoperation over a mobile internet connection by emulating control signals from the remote control receiver.

\subsection{Control Automation}
\label{ssec:method_control}
SAHA integrates several specialized controllers to ensure accurate and reliable operation in forest environments, consisting of a chassis controller for maintaining vehicle pose on uneven terrain, a driving controller for precise path following, and an arm controller for accurate control of SAHA's hydraulic arm.

\vspace{-5mm}
\subsubsection{Chassis Control}
\label{sssec:method_chassis_control}
With its compact design and short wheelbase, SAHA can navigate dense forests and operate in confined spaces.
However, this design also reduces stability on uneven ground, increasing the risk of tipping and limiting mobility on rough terrain.
To enhance stability, SAHA employs an active chassis with individually actuated legs and an active balancing control system that automatically maintains ground contact and keeps chassis orientation when traversing variable forrest terrain.

Our approach, similar to the work by Hutter~\etalcite{Hutter17ForceControl}, uses virtual model control to balance the chassis and cylinder force control to accurately track the desired forces on each leg.
A PID controller computes the virtual forces and torques required to achieve SAHA's target pitch, roll, and base height.
Then, we use hierarchical optimization to determine the optimal distribution of forces among the legs to generate the desired virtual forces and torques, subject to constraints on joint limits and leg contacts.


The cylinder force control is tested using two different hardware setups introduced in Sec.~\ref{ssec:method_hardware}.
For the setup using servo valves in the integrated control modules, the force command is sent directly to the module, which handles the low-level valve control internally.
For the setup using standard proportional valves, we implement a PID controller that regulates the cylinder pressure based on the desired force command and the measured pressure from the sensors.
The later setup, while less accurate and not capable of reaching the same bandwidth as the integrated control module, is more cost-effective and scalable.

\vspace{-5mm}
\subsubsection{Driving Control}
\label{sssec:method_driving_control}
The chassis balancing control on SAHA simplifies its driving control to a 2D \ac{CAV} control.
We use a kinematic model for center-articulated vehicles following the work by Corke~\etalcite{Corke01SteeringKinematics}.

\begin{figure}[t]
  \centering
  \includegraphics[width=\linewidth]{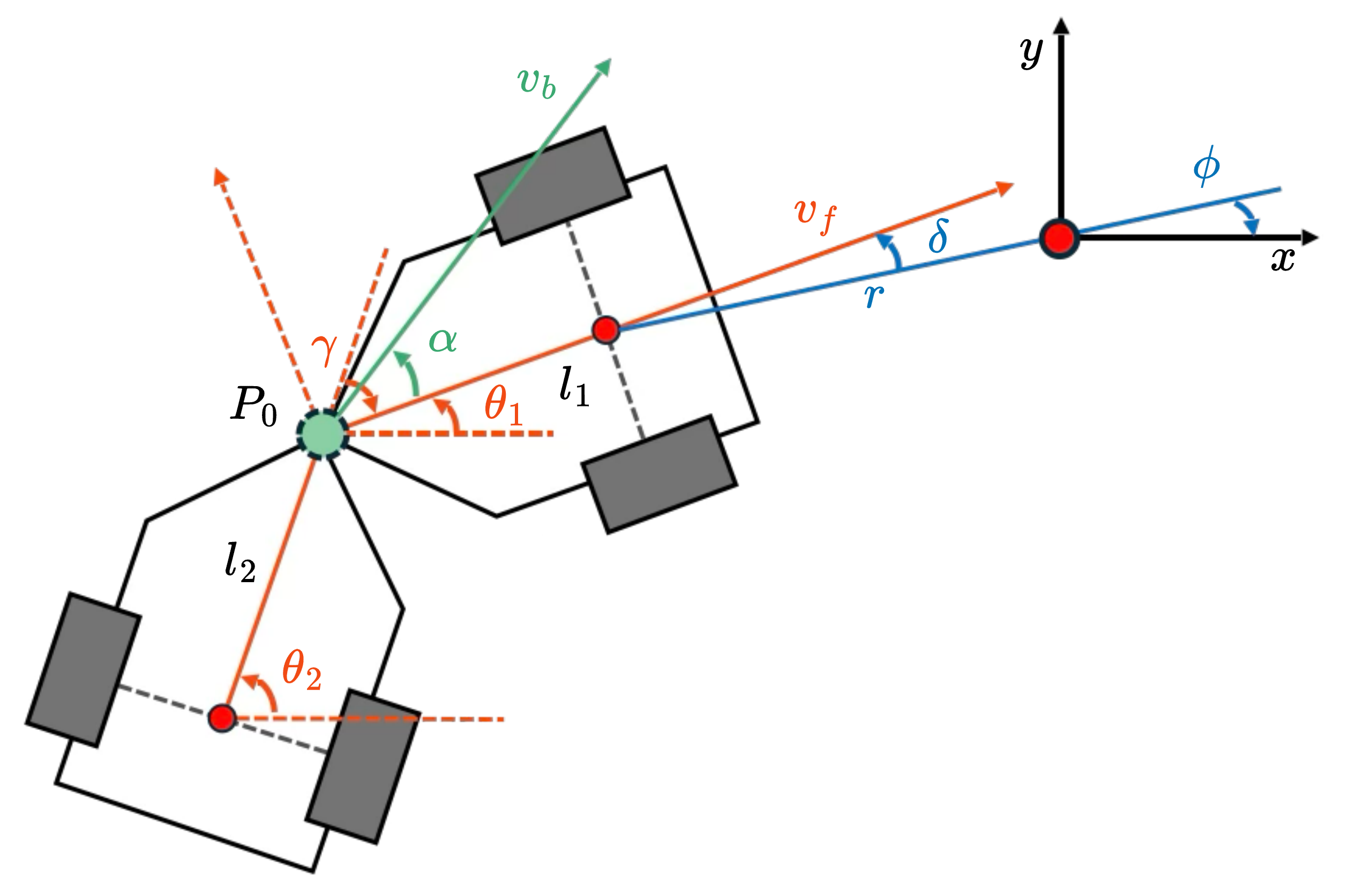}
  \caption{Kinematic model of the SAHA chassis as a \ac{CAV}. Adapted from~\cite{Hu24MotionPrimitives, Corke01SteeringKinematics}.}
  \label{fig:driving_control_model}
\end{figure}

As shown in~\cref{fig:driving_control_model}, the vehicle's relative pose with respect to a target point can be expressed in polar coordinates $[r, \phi, \delta]^{\mathsf{T}}$, where $r$ is the distance to the target, $\phi$ the direction to the target relative to the center of vehicle's front wheels, and $\delta$ the vehicle's heading relative to the direction to the target.
The vehicle's kinematics can be expressed as:
\begin{equation}
\begin{bmatrix} \dot{r} \\ \dot{\phi} \\ \dot{\delta} \end{bmatrix} =
\begin{bmatrix} -\cos\delta & 0 \\ \frac{\sin\delta}{r} & 0 \\ \frac{\sin\delta}{r} & 1 \end{bmatrix}
\begin{bmatrix} v \\ \dot{\theta}_1 \end{bmatrix},\label{eq:vehicle_kinematics}
\end{equation}
where $v$ is the vehicle's forward velocity, measured as the linear velocity of the articulation point, and $\dot{\theta}_1$ is the yaw rate of the front chassis, as depicted in~\cref{fig:driving_control_model}.

The model structure allows decomposition of the states into a slow subsystem $(r, \phi)$ and a fast subsystem $\delta$.
We then design a controller to stabilize the system with a backstepping approach.
Specifically, a stabilizing virtual control for the slow subsystem is designed as:
\begin{equation}
    \delta_{\text{ref}} = \arctan\left(k_\phi \phi\right),
\end{equation}
for some positive $k_\phi$.
Substituting $\delta = \delta_{\text{ref}}$ in Eq.~\eqref{eq:vehicle_kinematics} gives:
\begin{equation}
  \dot{r} = -\frac{v}{\sqrt{1 + k_\phi^2 \phi^2}}, \quad
  \dot{\phi} = -\frac{v k_\phi \phi}{r \sqrt{1 + k_\phi^2 \phi^2}},
\end{equation}
which suggests the slow subsystem asymptotically converges under some positive speed $v$.
To regulate $\delta$ to $\delta_{\text{ref}}$, we design the control law:
\begin{equation}
  \dot{\theta}_1 = \kappa v, \quad
  \kappa = -\frac{1}{r}k_\delta (\delta - \delta_{\text{ref}}) - \frac{\sin\delta}{r} - \dot{\delta}_{\text{ref}},
\end{equation}
with some positive $k_\delta$.

Because we only control the articulation angle $\gamma$ of SAHA, we use the following kinematic relation to compute the desired articulation speed~$\dot{\gamma}$:
\begin{equation}
  \dot{\gamma} = -\frac{l_1 + l_2}{l_2} \dot{\theta}_1 + \frac{\sin\gamma}{l_2} v,
\end{equation}
where $l_1$ and $l_2$ are the distances from the articulation point to the front and rear wheel axles, respectively, as shown in~\cref{fig:driving_control_model}.

While the system is stabilized independent of the speed $v$, the speed affects the convergence rate and the control effort.
We select $v$ heuristically based on the curvature $\kappa$ of the path to the target:
\begin{equation}
  v = v_{\text{max}} \frac{1-\beta |\kappa|^\lambda}{1-\beta |\kappa_{\text{max}}|^\lambda}.
\end{equation}

Intuitively, this encourages the vehicle to drive at maximum speed when it is driving straight to the target, and slow down when it needs to make sharp turns.

For a more detailed discussion of the path-following controller design, we refer the reader to the work by Hu~\etalcite{Hu24MotionPrimitives}.

\vspace{-5mm}
\subsubsection{Arm Control}
\label{sssec:method_arm_control}

\begin{figure}[t]
  \centering
  \includegraphics[width=\linewidth]{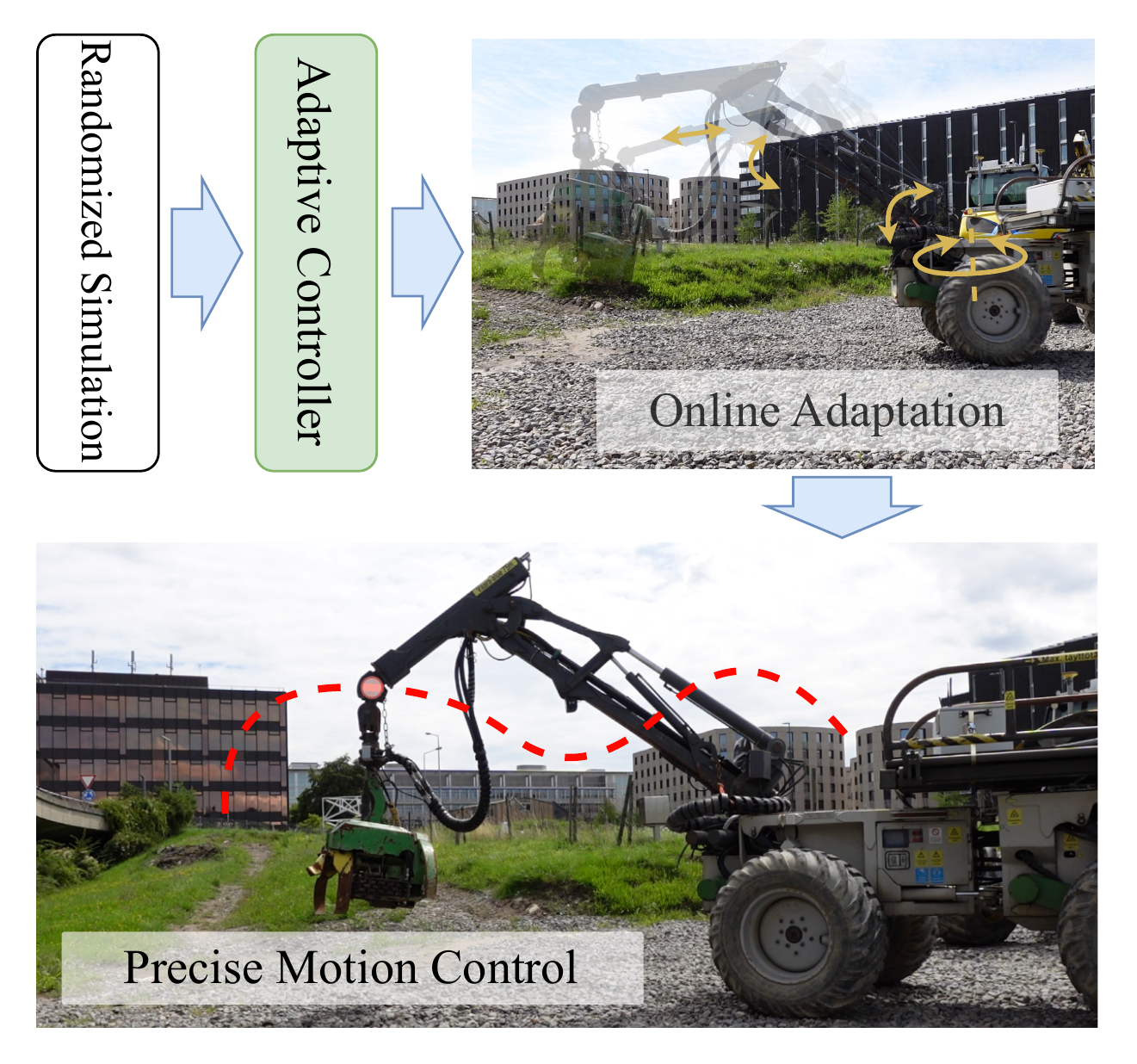}
  \caption{Diagram of the learning-based adaptive controller for hydraulic actuators used for SAHA arm control. Figure adapted from~\cite{Nan24LearningAdaptive}.}
  \label{fig:arm_control_diagram}
\end{figure}

We apply the learning-based adaptive controller developed by Nan~\etalcite{Nan24LearningAdaptive} for controlling the hydraulic arm of SAHA.
Unlike typical data-driven hydraulic controllers~\cite{Egli22GeneralApproach}, which require modeling specific systems, this approach uses a general hydraulic controller pretrained on a diverse set of simulated hydraulic systems and adapts to new, unseen systems with a small amount of real-world data.
This is achieved by training a latent-variable-conditioned neural network controller along with an energy-based model for latent parameter adaptation in a parameterized simulation environment, where the parameters are sampled from a prior distribution covering a wide range of hydraulic systems.
At deployment time, the controller infers latent variables from a small amount of data collected on the target system, allowing the controller to adapt to the specific dynamics of the new system.
\cref{fig:arm_control_diagram} illustrates the overall structure of the learning-based adaptive controller.
For details of the training and adaptation procedure, we refer the reader to the work by Nan~\etalcite{Nan24LearningAdaptive}.
We combine this learned hydraulic joint controller with an inverse kinematic controller, ensuring that the arm follows the desired trajectory while respecting the physical constraints of the system.

\subsection{Navigation}
\label{ssec:method_navigation}

\begin{figure}
  \centering
  \includegraphics[width=\linewidth]{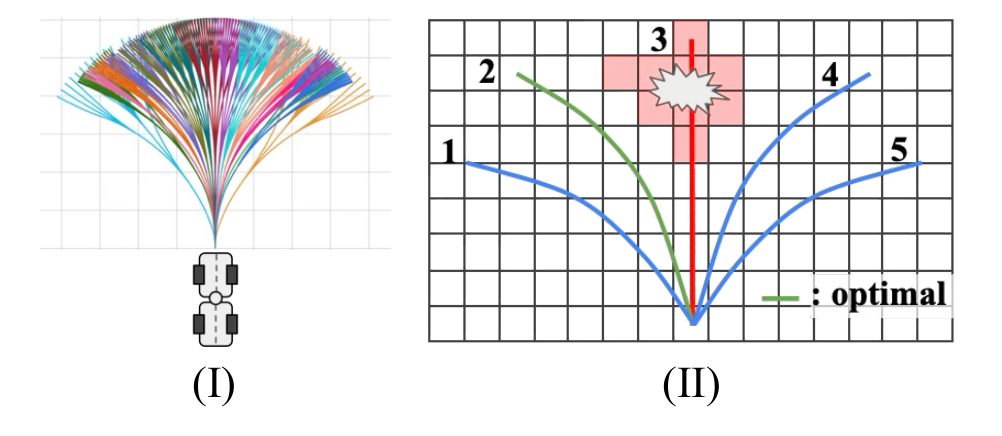}
  \vspace{-3mm}
  \caption{Visualization of the motion primitive-based planning approach for SAHA. (I) The motion primitives corresponding to a particular robot state are generated offline for a discretized set of steering angles and speeds. (II) The primitives are then checked online for collisions with the non-traversable obstacles in the environment, and the remaining primitives are evaluated with a cost function. The lowest-cost primitive is selected and executed. The figure is adapted from the work by Hu~\etalcite{Hu24MotionPrimitives}.}
  \label{fig:method_nav_primitive}
\end{figure}

For driving planning and obstacle avoidance, we adopt a receding-horizon planner built on motion primitives of a \ac{CAV}.
We first generate motion primitives offline using forward simulations of the \ac{CAV} kinematics model in a vehicle-centered frame.
We uniformly sample 30 discrete initial articulation angles $\gamma$ within the feasible region. For each initial state, we sample 15 control input pairs $(v_j,\dot\gamma_j)$ chosen within mechanical limits.
Each simulated trajectory splits twice, at $t=\tfrac{3}{v_j}$ and $t=\tfrac{6}{v_j}$, when a different steering command is sampled again.
We perform the forward simulation for a time horizon of $T_j=\tfrac{10}{v_j}$, yielding on average about 450 trajectories for each initial state.

During online planning, we combine the precomputed primitives with fast filtering. Potential collision points along every primitive are rasterized into grid cells using a simplified collision model of SAHA represented by three spheres.
Identified non-traversable obstacle points from a traversability map (see Sec.~\ref{ssec:method_perception_traversability}) are assigned to the grids, immediately culling occluded trajectories, as shown in~\cref{fig:method_nav_primitive}.
We then score the remaining candidates with a heuristic cost function accounting for progress towards the goal, heading direction alignment of both bodies, proximity to current steering state, nominal speed, terrain height, and short-horizon smoothness.
We finally select the primitive group with the lowest average cost.
A lookahead point from this group is passed to the driving controller as the immediate waypoint.

For further details on the motion primitive-based planner, we refer the reader to the work by Hu~\etalcite{Hu24MotionPrimitives}.

\vspace{-1mm}
\subsection{State Estimation}
\label{ssec:method_state_estimation}

We split our state estimation system into two components: an odometry module for pose-tracking and an initial localization module that locates SAHA within a prior global map.
The odometry component feeds into the traversability estimation and navigation systems, while the initial localization module enables targeting trees for the first thinning task that potentially lie beyond the robot's sensing range.

\vspace{-5mm}
\subsubsection{LiDAR-Inertial Odometry}
\label{ssec:odometry}

Our local traversability mapping system relies on the poses from the odometry when integrating traversability measurements into a map.
This traversability map is then used by the navigation system, which indirectly affects also the performance of the control systems.
It is therefore crucial that the odometry is robust and accurate to enable better downstream performance.
As our perception payload integrates a LiDAR and IMU (see Sec.~\ref{ssec:method_hardware}), we apply a recent open-source LiDAR-inertial odometry system, RKO-LIO~\cite{malladi2025arxiv}, for SAHA.
RKO-LIO offers a minimal set of tunable parameters and an out-of-the-box deployment capability on new platforms.
Furthermore, Malladi~\etalcite{malladi2025arxiv} report strong odometry performance on forestry data~\cite{malladi2025icra}, making it suitable for the SAHA robot which also has to operate in a forest.

For each LiDAR scan, the odometry performs scan-to-map point-to-point Iterative Closest Point (ICP) alignment.
The odometry maintains its own local map using the VDB data structure~\cite{museth2013siggraph}, which allows for efficient data association queries, and is updated after each new registered scan.
It uses a simple motion model, estimates a body acceleration using a Kalman filter, and uses an additional regularization on the ICP registration exploiting this body acceleration estimate.
The filter parameters and ICP regularization weights are adapted automatically from the IMU measurements, limiting additional parameter tuning.
For further details, we refer the reader to the work by Malladi~\etalcite{malladi2025arxiv}.

We evaluate the odometry's performance in the field and report the results in Sec.~\ref{sec:odometry_experiment}.
Although it could run well out-of-the-box on our onboard PC, we deploy the odometry on the Jetson Orin, which is a part of the perception payload.
This reduces the effects of network latency and ensures state estimates are available at the earliest possible time.
Since the Jetson Orin is considerably more resource-constrained than our onboard PC, to obtain odometry at sensor frame-rate, we limit the LiDAR range to 60\,m and also limit the number of points per voxel in the odometry map to 5.
These changes primarily reduce the computational load for ICP data association, which is challenging in forest environments due to numerous associations in the cluttered forest canopy.

\vspace{-6mm}
\subsubsection{Initial Localization in Prior Map}

\begin{figure}[t]
\centering
\includegraphics[width=\columnwidth]{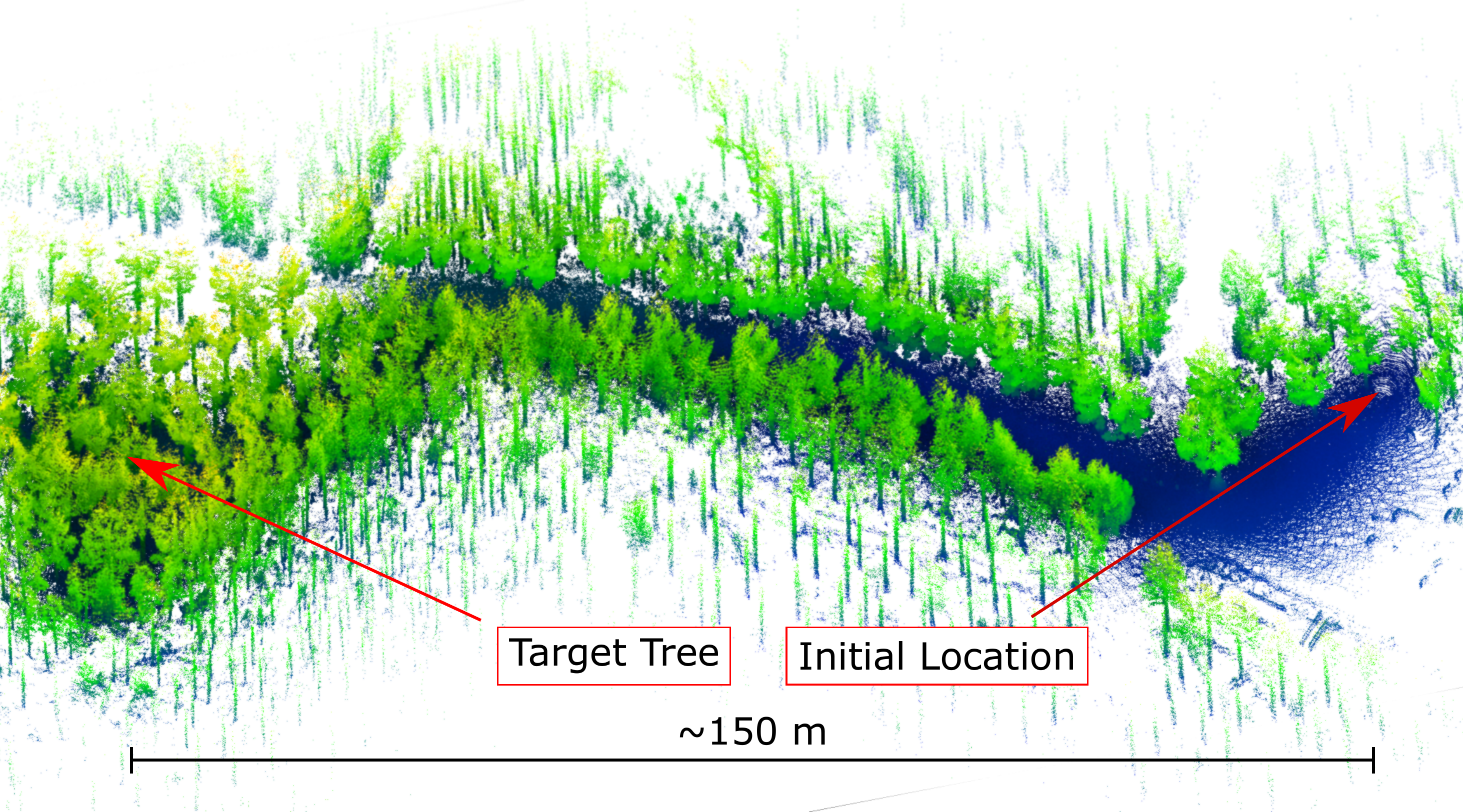}
\caption{Provided global map of the forest region where the SAHA was tested, generated from prior mapping. SAHA's initial location during a test is shown on the right, and the target tree for thinning is on the left, approximately 150\,m apart in straight-line distance.}
\label{fig:global_localization}
\vspace{-3mm}
\end{figure}

While the odometry module provides accurate local pose tracking, SAHA must initially locate itself within a global map frame in order to plan routes to trees beyond sensing range.
\cref{fig:global_localization} shows a map we used in one of SAHA's missions, along with the location of SAHA and the target tree for thinning during one of our tests.
The map is generated offline using data from prior forest mapping with a backpack-mounted sensor rig, following a method similar to the work of Malladi~\etalcite{malladi2025icra}.

For initial localization in this map, we use an approach similar to the odometry's LiDAR registration step~\cite{malladi2025arxiv}, replacing the online odometry map with the pre-built global map.
We assume the robot remains stationary during localization and an initial pose guess can be provided, so that a subsequent LiDAR scan can be registered against the global map.
We voxelize the global map, \(\mathcal{M}\), using a VDB volume, leveraging its spatial indexing capabilities~\cite{museth2013siggraph} to reduce the number of points processed and also accelerate the nearest-neighbor search during data association.
For each point \(\mathbf{s}\) in the scan~\(\mathcal{S}\), we find the nearest neighbor \(\mathbf{m}\in\mathcal{M}\) within a fixed radius threshold to form a set of correspondences \(\mathcal{C}=\{(\mathbf{s},\mathbf{m})\mid \mathbf{s}\in\mathcal{S},\;\mathbf{m}\in\mathcal{M}\}\).
The forest surroundings may have changed from the mapping to the thinning mission, e.g., trees were logged or vehicles were parked nearby.
To mitigate the effect of such outliers in the registration, we employ a robust kernel and minimize the point-to-point residual
\begin{equation}
\chi(\mathbf{T})
  = \sum_{(\mathbf{s},\mathbf{m})\in\mathcal{C}}
      \rho\bigl(\|\mathbf{T}\,\mathbf{s}-\mathbf{m}\|^{2}\bigr),
\end{equation}
where $\mathbf{T} \in \mathrm{SE}(3)$ is the current pose estimate and \(\rho\) is the Geman-McClure robust kernel.
The iterative least-squares solution is obtained by solving
\begin{equation}
\begin{split}
\Delta\boldsymbol{x}^{\star}
  &= \arg\min_{\Delta\boldsymbol{x}\in\mathbb{R}^{6}}
    \chi\bigl(\mathbf{T}\boxplus\Delta\boldsymbol{x}\bigr), \\
\mathbf{T} &\leftarrow \mathbf{T}\boxplus\Delta\boldsymbol{x}^{\star},
\end{split}
\end{equation}
until convergence, where \(\Delta\boldsymbol{x}\in\mathbb{R}^{6}\) is a correction vector and \(\boxplus\) applies this vector to the pose estimate~\cite{hertzberg2013if}.

Once the global pose is estimated, the target tree location from the mission plan can be located in SAHA's odometry frame.
The odometry subsystem starts pose-tracking, with the traversability system using the estimated pose to build a local traversability map.
The navigation module then uses this map to plan a safe path to the tree.

\vspace{-2mm}
\subsection{Traversability Mapping}
\label{ssec:method_perception_traversability}

For safe navigation in the forest, the SAHA robot needs to be aware of traversable terrain.
We estimate a per-point traversability classification from each LiDAR scan, which we subsequently integrate into a local traversability map in a probabilistic manner.

\vspace{-6mm}
\subsubsection{Traversability Classification}
\label{ssec:method_perception_traversability_classification}

\begin{figure}[t]
  \centering
  \includegraphics[width=\columnwidth]{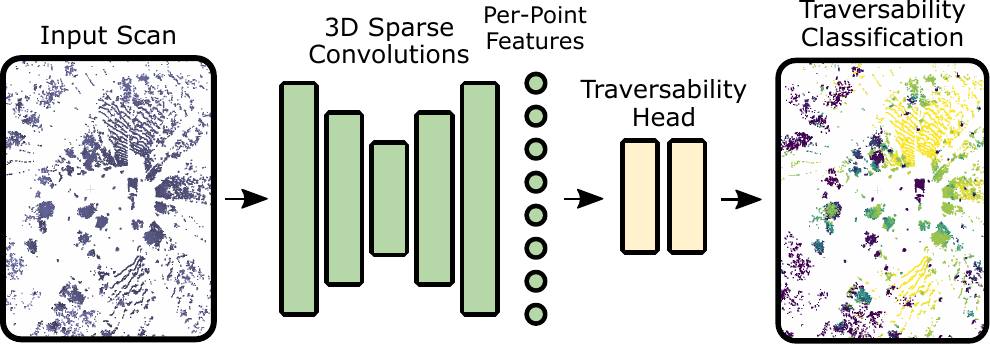}
  \caption{Overview of the traversability classification network architecture. We process the input LiDAR scan using a sparse 3D convolution backend and a traversability head, producing a per-point traversability score.}
  \label{fig:traversability_classification_network_architecture}
  \vspace{-5mm}
\end{figure}

We classify each scan from the Hesai XT32 sensor on our perception payload using a deep learning model that estimates a per-point traversability score.
The architecture of this model is illustrated in~\cref{fig:traversability_classification_network_architecture}.

To process the 3D LiDAR data, we use a Minkowski engine-based MinkUNet backbone~\cite{choy2019cvpr} with 8~layers and 205k~parameters.
Using this backbone, we produce a per-point feature embedding of dimension~16.
We then pass this embedding into a multi-layer perceptron, referred to as the traversability head, which outputs a single traversability score between~0~for non-traversable and~1~for traversable.
The traversability head is a small network with two hidden layers of the same size as the embedding.
Except the last layer, which uses a sigmoid activation, all other layers use ReLU activation.
With a total of 205.6k~trainable parameters, the result is a compact network capable of both fast training and efficient inference.

\begin{figure}[t]
  \centering
  \includegraphics[width=\columnwidth]{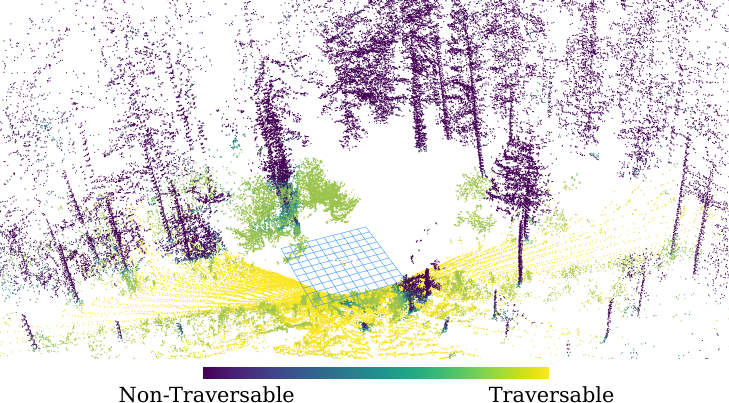}
  \caption{Output of the traversability classification network on a single scan from the DigiForests dataset. Points are colored according to a heatmap of the predicted traversability score, where higher values (yellow) indicate traversable regions and lower values (purple) indicate non-traversable regions.}
  \label{fig:traversability_classification_single_scan}
  \vspace{-2mm}
\end{figure}

To supervise the training of this network, we avoid direct labeling of data from the testing region.
Instead, we leverage an open-source dataset designed for panoptic segmentation in forest environments: the DigiForests dataset~\cite{malladi2025icra}.
This dataset contains data collected in forests in Switzerland using backpack-mounted sensor payloads recorded across three different seasons. Recordings in each season use a different LiDAR sensor configuration.
Although the backpack-mounted sensor configurations differ from SAHA's perception payload, we found that our model trained on DigiForests data could still generalize well to our target application.
DigiForests provides semantic annotations for four classes: ground, shrub, tree stem and tree canopy.
We remap these classes each to a traversability score, assigning 1~for ground and 0~for tree stem and canopy.
The SAHA robot can indeed traverse small bushes or undergrowth in the forest.
However, the dataset provides only a general shrub class representing undergrowth of varying density and size.
Therefore, we assign a score of 0.8~to shrubs, neither marking them as fully traversable nor discarding them entirely as non-traversable.
This design choice allows us to later fine-tune the score threshold for deciding traversable regions and adjust the mapping behaviour during deployment.
In our evaluations, we observe that the network successfully learns to distinguish between ground and shrub regions, as shown in~\cref{fig:traversability_classification_single_scan}.

\vspace{-6mm}
\subsubsection{Probabilistic Mapping}
\label{sec:probabilistic_mapping}

Separate from the map used for odometry, we integrate each traversability-classified LiDAR scan into a local traversability map.
We maintain this map at a higher spatial resolution than the odometry map, making it suitable for use in the navigation module.
Conversely, we reduce the spatial extent of the map to ensure the perception system operates at sensor frame rate.
For each voxel, we maintain a belief of both occupancy and traversability, as there can be objects that SAHA can drive over, i.e., occupancy does not imply non-traversability.
Occupancy mapping allows us to handle sensor noise as well as dynamic obstacles, which should not be included in the static traversability map.
Furthermore, given the traversability classification model operates at sensor frame rate, accumulating independent predictions over time effectively reduces prediction errors~\cite{mersch2023ral}.
This is particularly valuable since we rely on our model's generalization capabilities and do not fine-tune it with labeled data from the deployed forest or sensor setup.

For occupancy mapping, we model the occupancy state of each voxel as a log-odds value, representing the belief in whether the voxel is occupied or free.
Each point in a LiDAR scan increases the occupancy log-odds of the corresponding voxel by a predefined increment.
To model free space, we perform ray casting from the sensor origin to each detected point. We once again leverage the VDB data structure~\cite{museth2013siggraph} for our traversability map, which enables efficient ray traversal through its cache-coherent access patterns.
Voxels traversed by a ray but not containing a detected point are updated as free by reducing their occupancy log-odds.
We maintain a log-odds belief for traversability in a similar manner.
Since our traversability network produces a score for each point in the scan, we integrate the score into the map independently of the occupancy state.
For each voxel corresponding to a classified point, we use the logit from the final layer of the traversability network and accumulate the information over time, similar to the approach of Mersch~\etalcite{mersch2023ral}.

When querying the map for occupied voxels and their traversability probabilities, we first collect voxels with an occupancy probability greater than 0.5.
We then recover the posterior traversability probability $p(x)$ for those occupied voxels using the stored traversability log-odds $l(x)$ as
\begin{equation}
  p(x) = 1 - \frac{1}{1 + \mathrm{exp}\,l(x)}. \label{eq:logodds_to_prob}
\end{equation}

\begin{figure}[t]
  \centering
  \includegraphics[width=0.9\columnwidth]{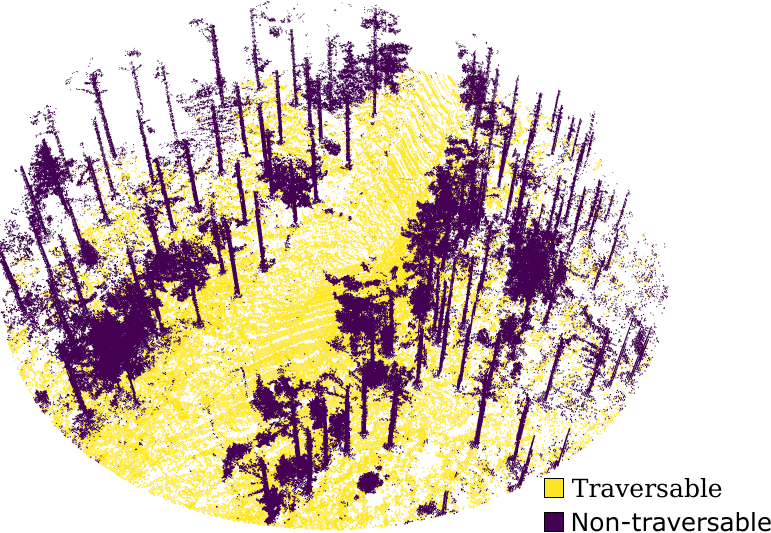}
  \caption{Qualitative result of the traversability mapping pipeline on the SAHA robot. Points in yellow are considered traversable, with traversability probability higher than~0.8, whereas points in purple are non-traversable, with a lower probability than the threshold.}
  \label{fig:traversability_mapping_saha_qualitative}
\end{figure}

A qualitative result of our traversability mapping system is shown in~\cref{fig:traversability_mapping_saha_qualitative}, in which we consider a voxel as traversable if $p(x) > 0.8$.

\section{Experimental Results}


We perform a series of experiments to evaluate the individual components of the SAHA system as well as the fully integrated system in real forest environments.
The results are presented in the following order.
First, we detail the results of experiments evaluating individual components: LiDAR-inertial odometry, traversability classification, local path planning, and control modules for balancing, driving, and arm manipulation.
Subsequently, we report field trials conducted in real forest environments between fall 2024 and summer 2025, showing the operational effectiveness of SAHA.

\subsection{State Estimation}
\label{sec:odometry_experiment}

\begin{figure}
  \centering
  \includegraphics[width=\linewidth]{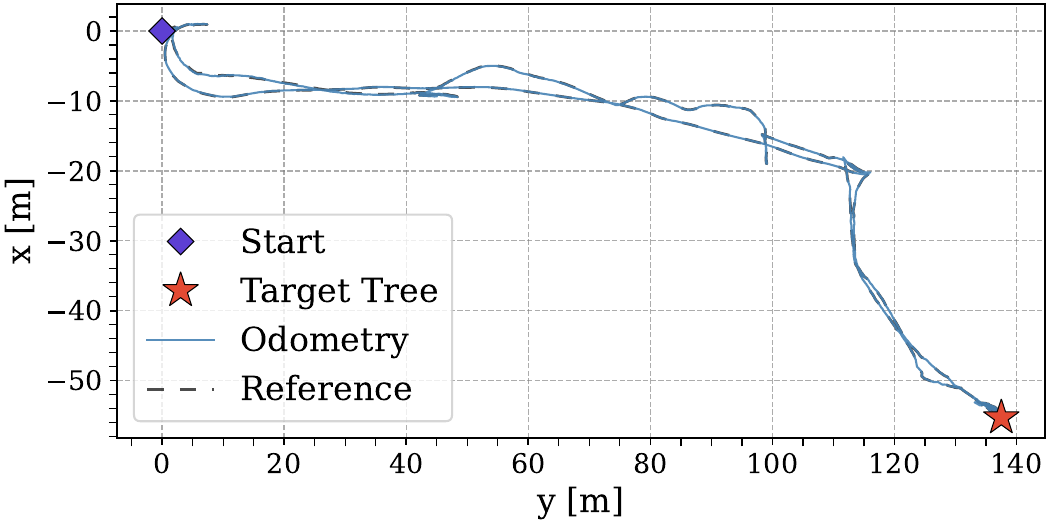}
  \caption{Estimated odometry (blue) and reference trajectory (dashed) from an autonomous thinning mission lasting 27\,min. The reference was obtained through offline LiDAR bundle adjustment. SAHA begins near the top left and travels toward the target tree at the bottom right. After approaching the target tree, SAHA returns to the starting position.}
  \label{fig:saha_rko_lio_result}
  \vspace{3mm}
\end{figure}

The first experiment evaluates the performance of the odometry subsystem.
Previous work~\cite{malladi2025arxiv} has demonstrated the performance of the odometry on data from multiple different platforms and environments.
In this work, we test this odometry's performance in real-world forests in the context of autonomous forest thinning.
To assess odometry accuracy, we analyze the results from two real-world field tests: one where the robot was operational for 27\,min covering 528.5\,m, and another lasting 76\,min over 1.01\,km.
After the field experiments, we generated reference trajectories using an offline LiDAR bundle adjustment system~\cite{wiesmann2024arxiv-ba}.
A qualitative comparison of the odometry estimate and the reference trajectory for one of the experiments is shown in~\cref{fig:saha_rko_lio_result}.

For quantitatively evaluating odometry accuracy, we use two widely adopted metrics.
The first is absolute trajectory error (ATE) after alignment, which provides a measure of global drift in the estimated trajectories.
The second is the relative pose error (RPE), which measures the average translational error between estimated and reference trajectories over various segment lengths, reported as a percentage.
This approach is common in standard benchmarks~\cite{geiger2012cvpr}, though we use shorter intervals of 1, 2, 5, 10, 20, 50, and 100\,m to better match the scale of our experiments in a forest~\cite{guadagnino2025icra}.
Across the two experiments, the average RPE of the odometry was 3.35\% and average ATE was just 6.34\,cm.
The odometry overall exhibited very low drift, as indicated by the low RPE and especially ATE results.
This can also be seen qualitatively in~\cref{fig:saha_rko_lio_result}.

\begin{figure}[t]
  \centering
  \includegraphics[width=\linewidth]{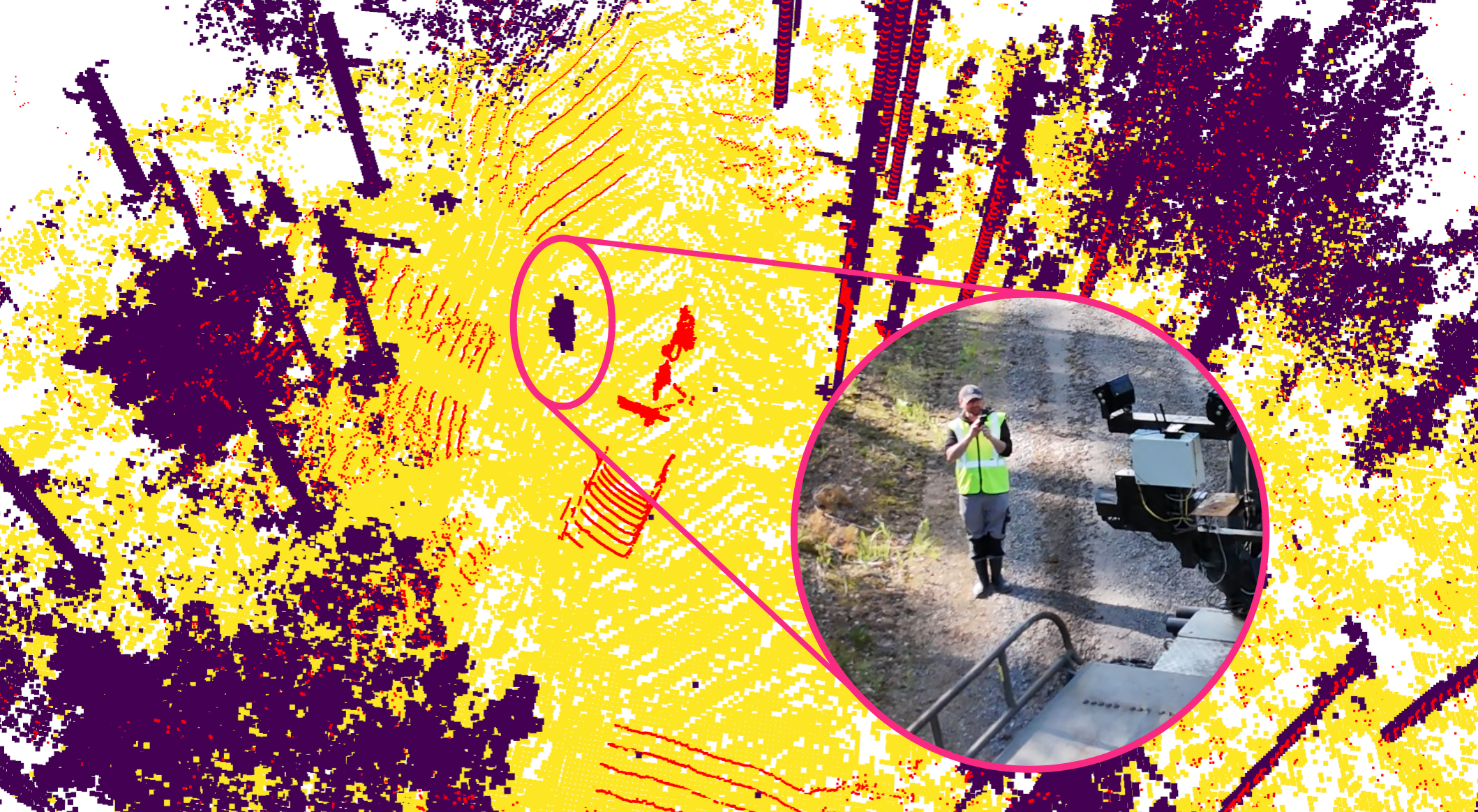}
  \caption{Traversability map generated during autonomous SAHA operation. Points in yellow are considered traversable and points in purple are non-traversable. The raw LiDAR scan is overlaid in red. A human standing near the robot is highlighted in the circle, who is classified as non-traversable in the map. The training data for the traversability classification model does not include annotations for humans.}
  \label{fig:saha_traversability_human}
\end{figure}

\subsection{Traversability Classification}

In this experiment, we evaluate the performance of our traversability classification approach.
As detailed in Sec.~\ref{ssec:method_perception_traversability}, we did not label any data manually for training our deep learning model.
We trained our model using the DigiForests dataset~\cite{malladi2025icra} which provides semantic and instance annotations.
Following earlier work~\cite{guan2023ar}, we convert the ground truth semantic labels to either 0 or 1 to indicate traversability, similar to how we trained the model.
However, unlike for training, for evaluation we consider shrubs to also be non-traversable and assign it a score of 0 to simplify evaluation~\cite{guan2023ar}.
Furthermore, this reflects the behaviour when we deploy the system in full, as we threshold the traversability probability queried from the probabilistic map before using the path planner.
On the validation split of the DigiForests dataset, the model then achieves 76.62\% accuracy for the traversable regions and 94.76\% accuracy for the non-traversable regions, with a mean accuracy of~85.69\%.

\begin{figure*}[t]
  \centering
  \subfigure[]{
    \centering
    \includegraphics[width=0.4\linewidth]{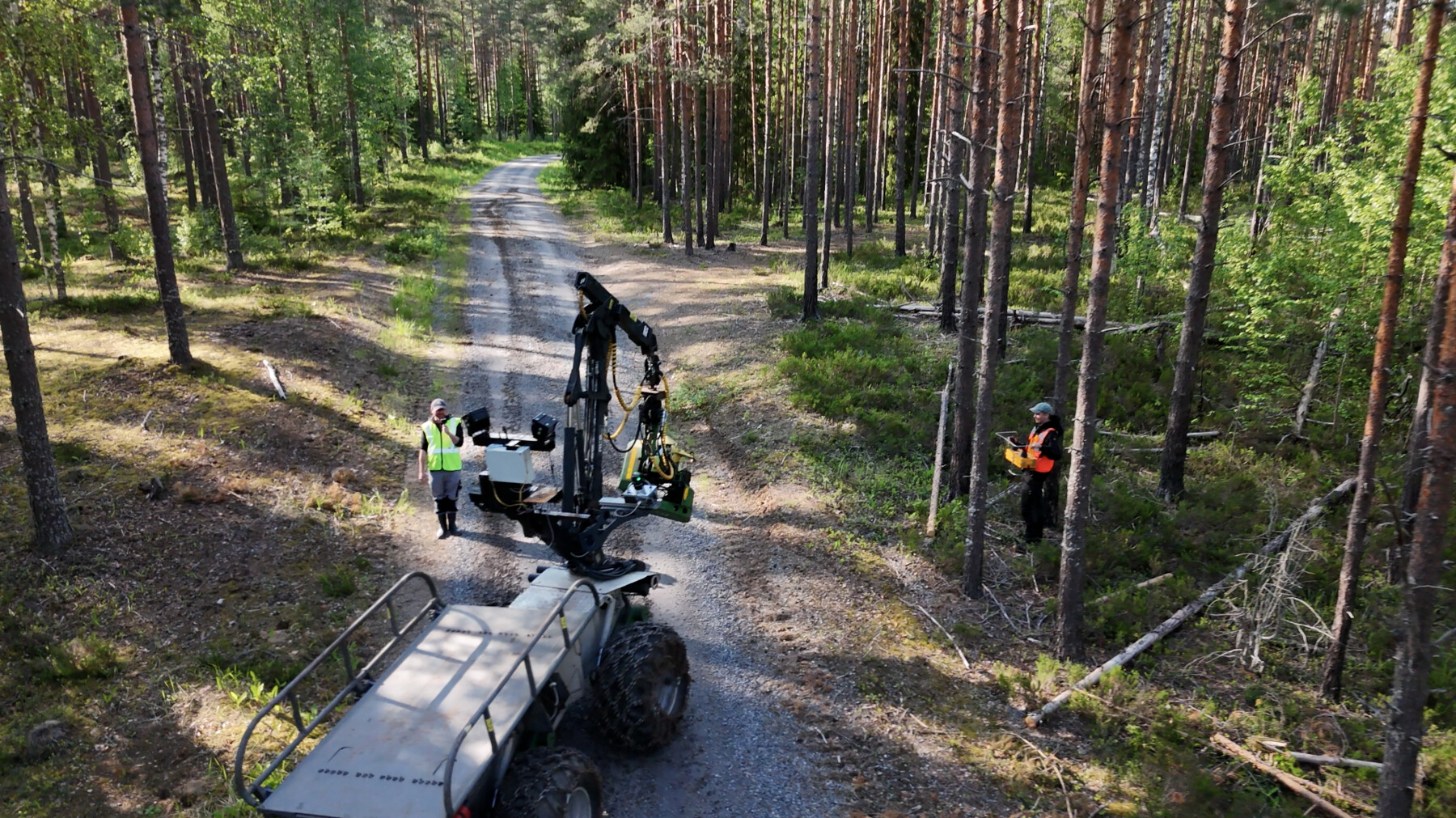}
  }
  \subfigure[]{
    \centering
    \includegraphics[width=0.4\linewidth]{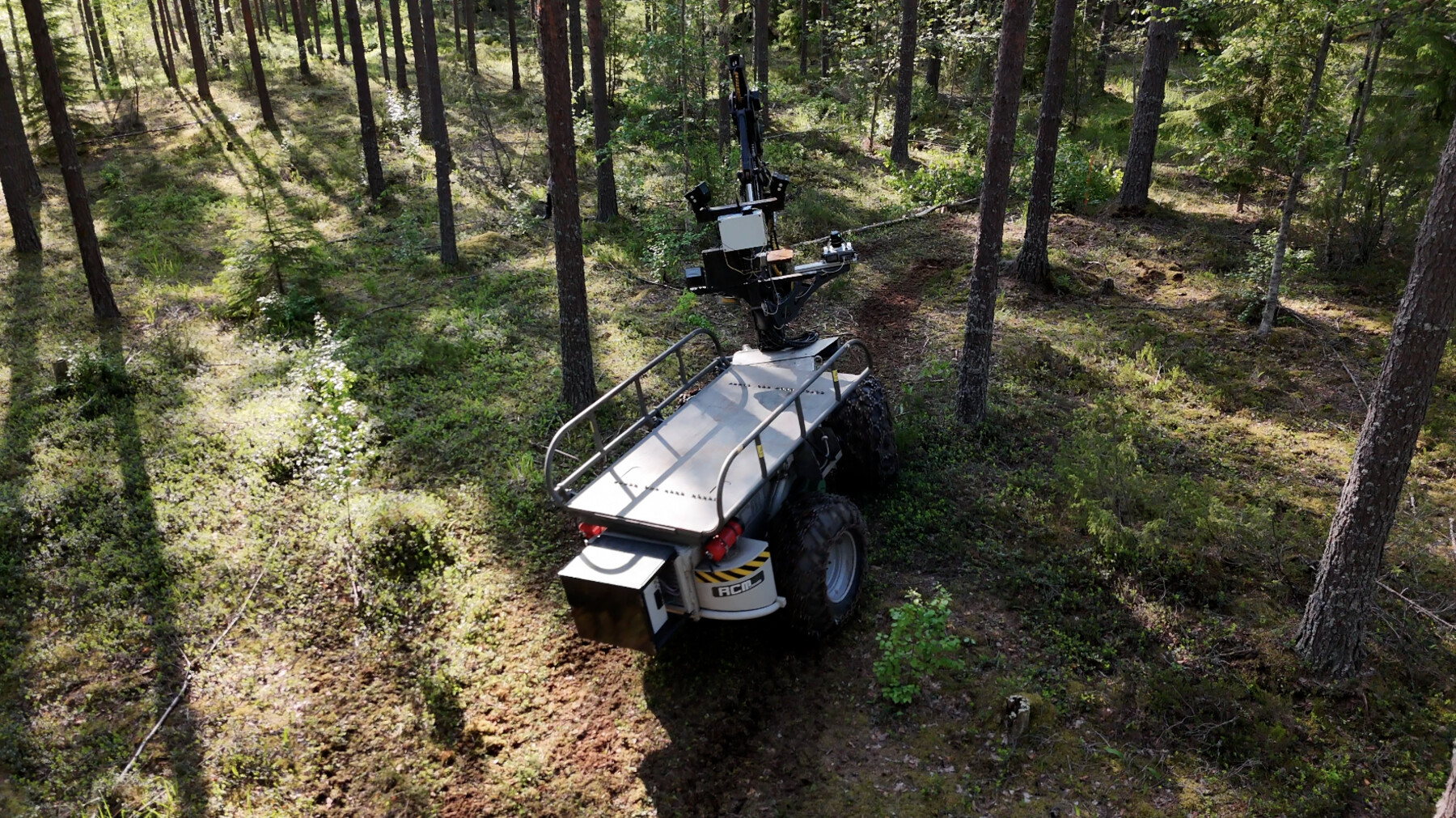}
  }\\
  \subfigure[]{
    \centering
    \includegraphics[trim={8cm 0cm 8cm 11cm}, clip, width=0.4\linewidth]{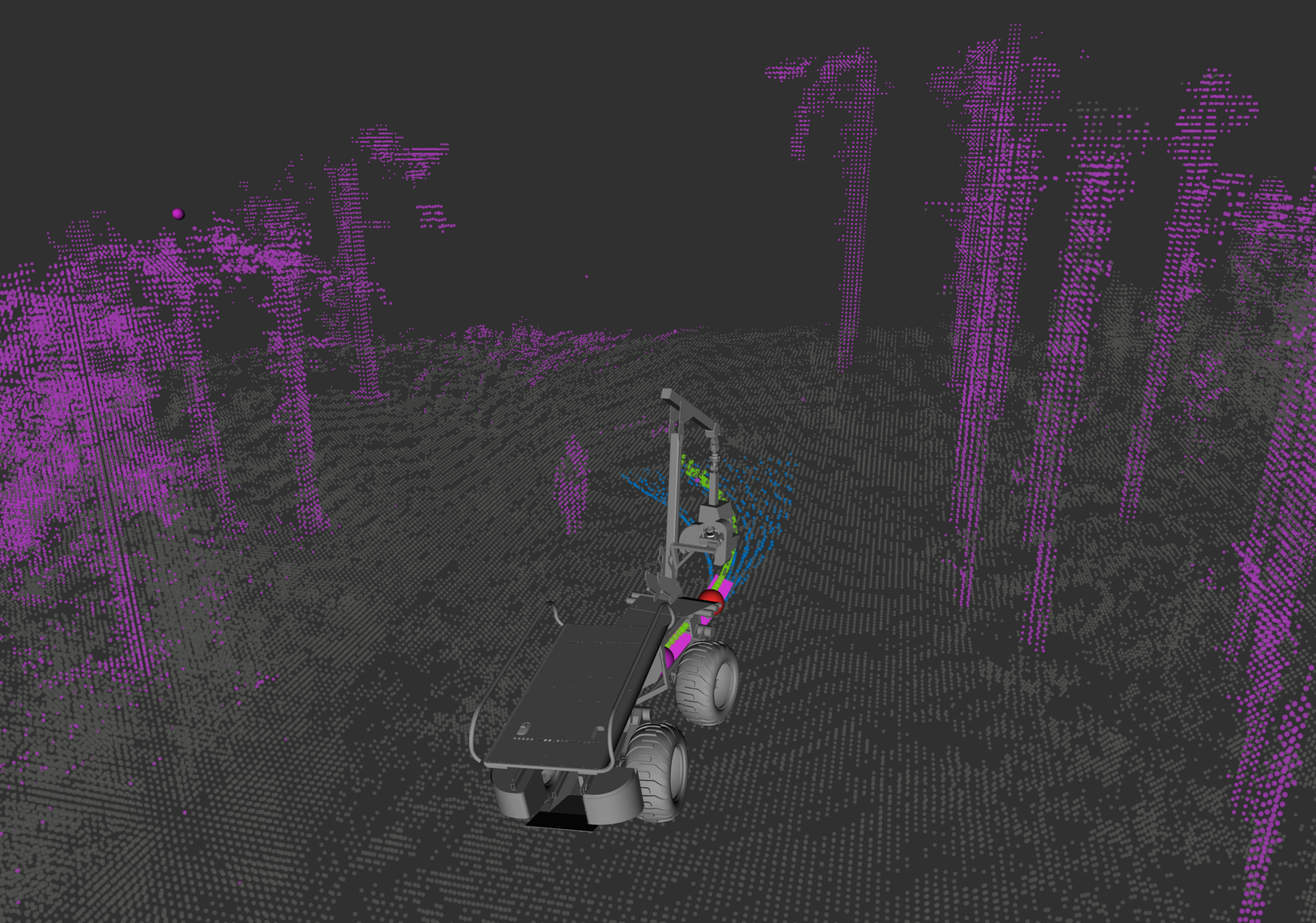}
  }
  \subfigure[]{
    \centering
    \includegraphics[trim={8cm 0cm 8cm 11cm}, clip, width=0.4\linewidth]{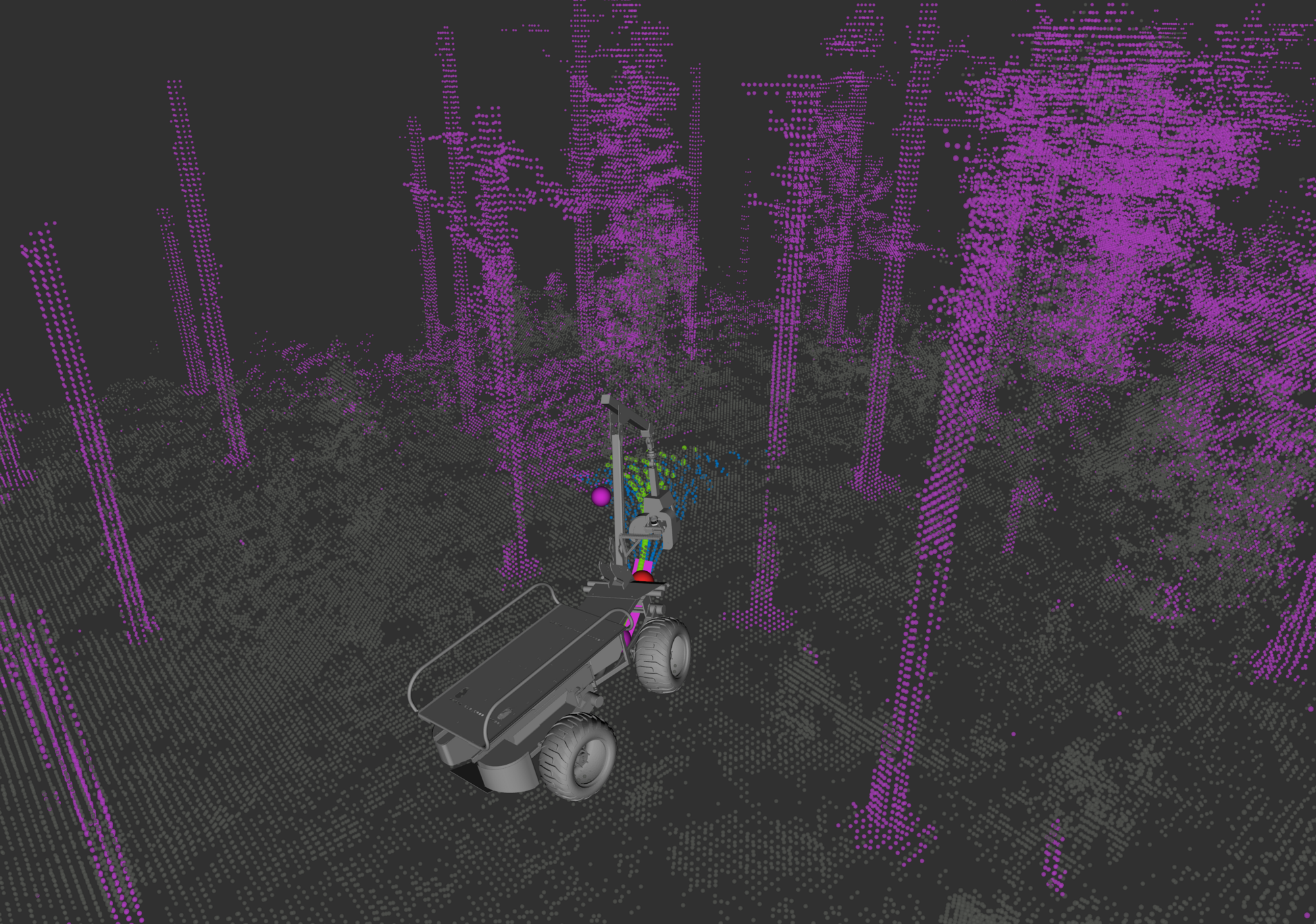}
  }
  \vspace{3mm}
  \caption{Pictures during SAHA navigating toward a target tree from a distance. (a) Aerial view of SAHA avoiding a human while driving on a forest road toward a target tree. (b) Aerial view of SAHA driving through dense trees toward a target tree. (c)(d) Corresponding planning result visualizations, in which the perceived point cloud is classified into traversable (gray) and non-traversable (pink). The motion primitives after collision checking are shown in blue, and the selected primitive group is highlighted in green. The purple sphere in (d) indicates the target tree.}
  \label{fig:saha_planning_result_qualitative}
  \vspace{2mm}
\end{figure*}

In~\cref{fig:saha_traversability_human}, we show a qualitative result of the traversability map produced during autonomous robot operation.
The figure highlights a person standing in front of SAHA.
This is a challenging condition for the approach, as the training data does not include annotations for humans.
Nevertheless, the person is still classified as non-traversable in the map.
The planner then uses this map information to execute an evasive manoeuvre, avoiding the person as shown in~\cref{fig:saha_planning_result_qualitative}.
We note again that we did not fine-tune the model with data from the target sensor or forest, which indicates the approach's generalization capability for traversability classification.

\vspace{-5mm}
\subsection{Local planning}

\begin{figure}[t]
  \centering
  \includegraphics[width=\linewidth]{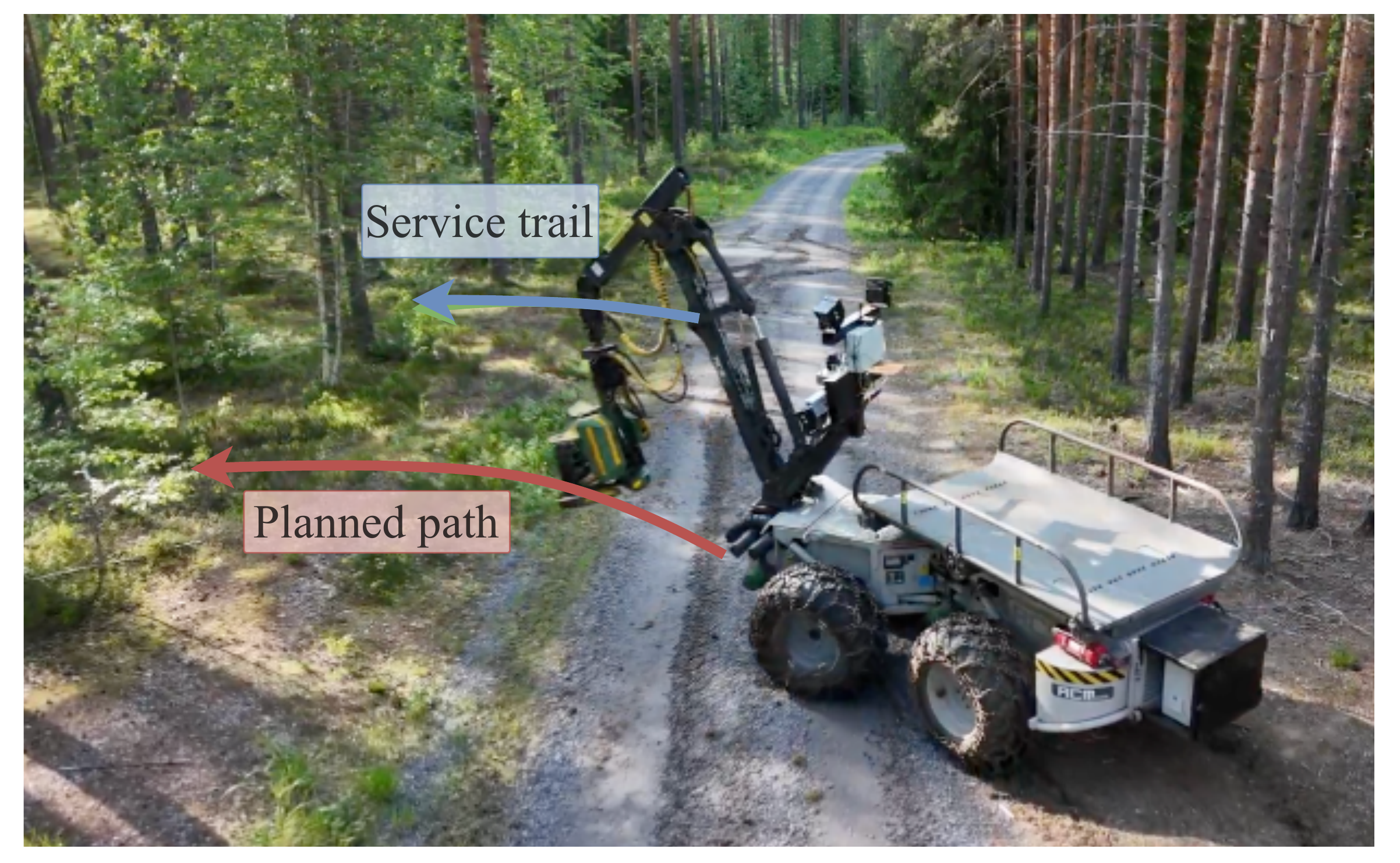}
  \vspace{-2mm}
  \caption{A failure case of the planner. When the target tree is located within the forest on the left side of the road, the planner attempts to enter the forest whenever it detects a gap between trees, instead of following the intended service trails.}
  \label{fig:planner_failure}
  \vspace{-2mm}
\end{figure}

The next experiment qualitatively evaluates the local planner within real forest environments.
Previous work~\cite{Hu24MotionPrimitives} has demonstrated the effectiveness of the motion primitive-based local planner in simulation, along with short field tests in open space with artificial obstacles.
In this work, we further evaluate the local planner in extensive experiments in real forest environments and within the presented autonomous thinning pipeline.
We deployed the planner in full cutting missions involving navigation on forest roads and service trails to reach target trees, and the planner demonstrated reliable performance in all experiments.
\cref{fig:saha_planning_result_qualitative} shows qualitative results of the planner in two different scenarios: avoiding a person while driving on a forest road and negotiating turns through trees to reach a target.
In both cases, the planner successfully generates safe trajectories avoiding non-traversable obstacles while driving towards the destination.

While the primitive-based planner does not rely on a prior map of the environment, an unavoidable limitation is that it attempts to drive wherever it detects a traversable path.
In real-world forest applications, this can lead to suboptimal behavior, as it is generally preferable to follow existing service trails created by previous runs of forestry machines in order to minimize environmental impact.
As shown in~\cref{fig:planner_failure}, when a target tree is located inside a forest, the planner attempts to enter the forest whenever a traversable gap is wide enough, rather than following the service trail, which may be only a few meters away.

\subsection{Control}
The evaluation of the control system includes the three main components, balancing control, driving control, and arm manipulation control.
All modules were individually evaluated first, either in simulation or in the real world, before being integrated in the SAHA system.

\subsubsection{Balancing Controller}
\begin{figure}
  \centering
  \includegraphics[width=\linewidth]{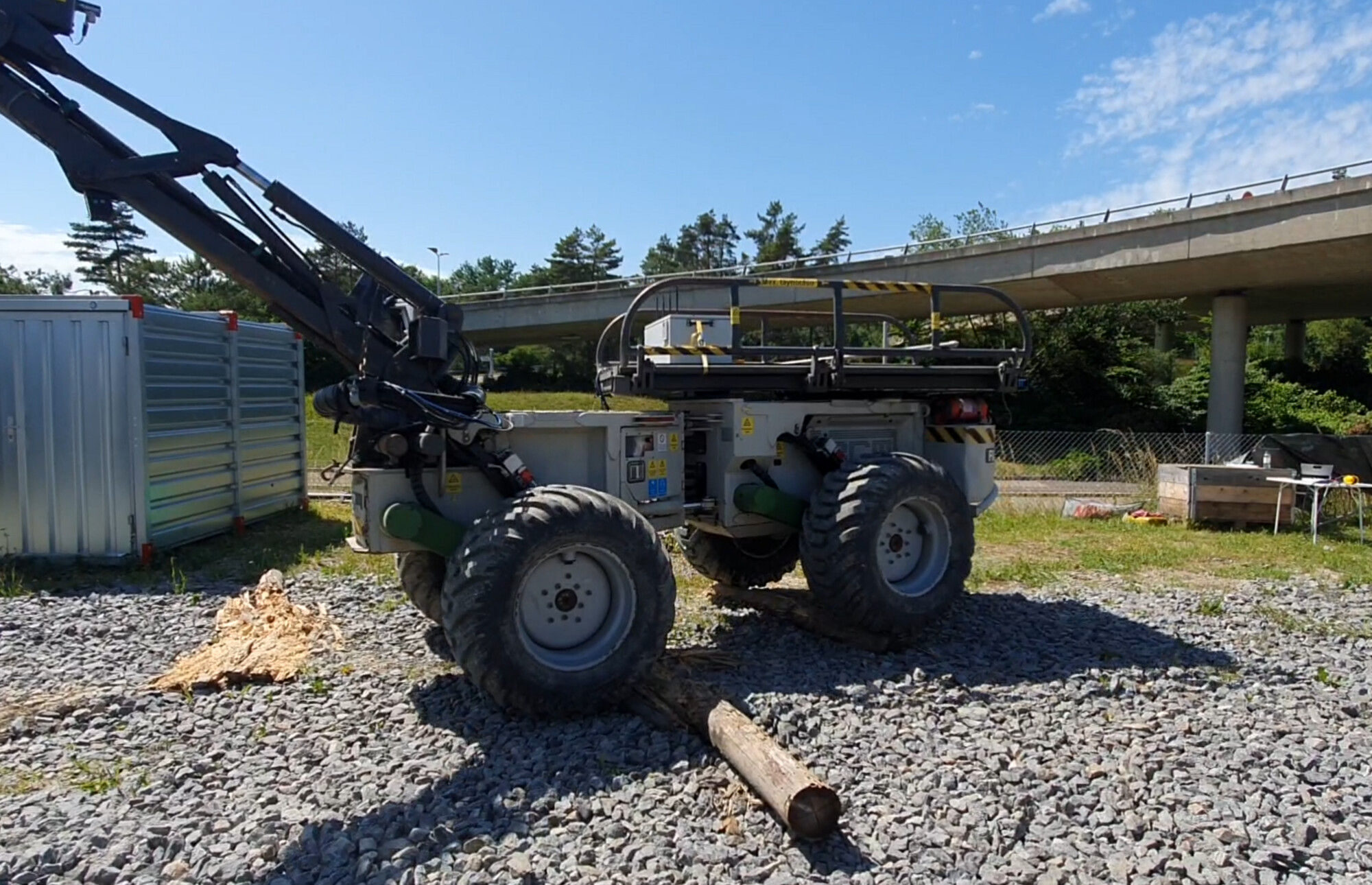}
  \caption{SAHA, equipped with integrated control modules on its chassis joints, balances its chassis while driving over logs on a test site.}
  \label{fig:saha1_over_logs}
\end{figure}

\begin{figure}
  \centering
  \subfigure[Cylinder forces during balancing test.]{
    \centering
    \includegraphics[width=\linewidth]{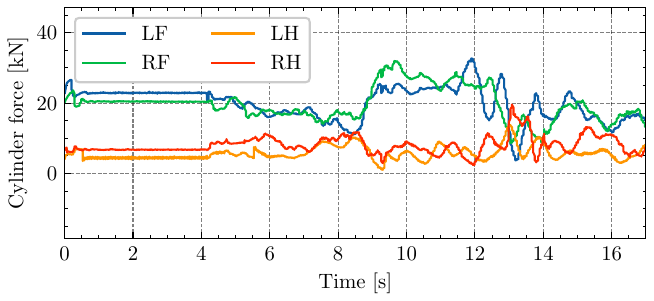}
    \label{fig:balancing_forces}
  }
  \subfigure[Roll and pitch during balancing test.]{
    \centering
    \includegraphics[width=\linewidth]{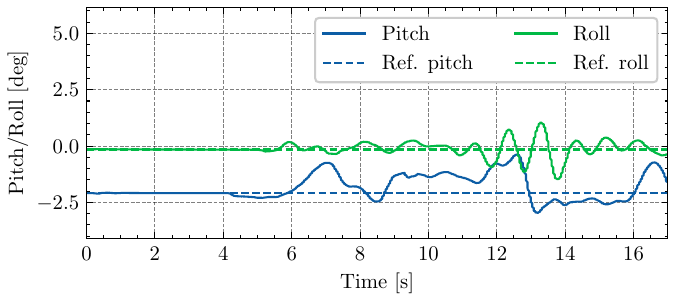}
    \label{fig:balancing_pose}
  }
  \caption{Balancing controller performance evaluated on the real robot.}
  \label{fig:balancing_plots}
\end{figure}

\begin{figure}
  \centering
  \includegraphics[width=\linewidth]{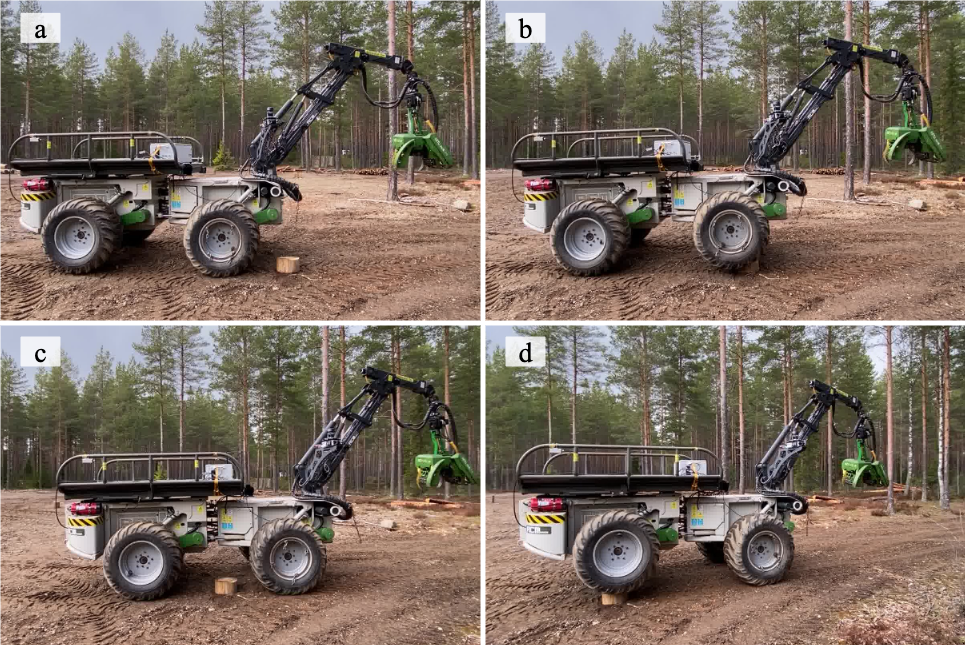}
  \caption{SAHA, equipped with proportional valves, balances its chassis when driving over a stump on a test site.}
  \label{fig:saha2_over_stump}
\end{figure}

The balancing controller was evaluated on the real robot by driving over obstacles at a test site, as shown in~\cref{fig:saha1_over_logs} and~\cref{fig:saha2_over_stump}.
We tested two different approaches for balancing control, as detailed in Sec.~\ref{sssec:method_chassis_control}.
\cref{fig:saha1_over_logs} shows a test with the integrated control modules for chassis balancing.
The balancing controller is enabled when the SAHA stands still in front of the log, then an operator commands the robot to drive forward over it.
The leg cylinder forces and the chassis roll and pitch during driving are shown in~\cref{fig:balancing_plots}.
\cref{fig:balancing_pose} demonstrates that SAHA's chassis roll and pitch remain close to the desired positions, with deviations below 2°.
This stability is achieved by continuously adjusting the leg cylinder forces, as shown in~\cref{fig:balancing_forces}.
If the chassis balancing is disabled, SAHA reaches a maximum tilt of 4.41° in pitch and 5.6° in roll directions to go over the same obstacle, and the diagonal rear wheel loses contact with the ground in the process.

The same experiment was repeated with standard proportional valves to actuate the leg cylinders instead of the high-bandwidth servo valves in the integrated control modules.
As shown in~\cref{fig:saha2_over_stump}, the balancing controller was still able to keep the chassis balanced while driving over a stump.
Although larger deviations in roll and pitch were observed compared to the previous test, the balancing controller still succeeded in keeping all four legs in contact with the ground.

\vspace{-5mm}
\subsubsection{Driving Controller}
\begin{figure}
  \centering
  \includegraphics[width=\linewidth]{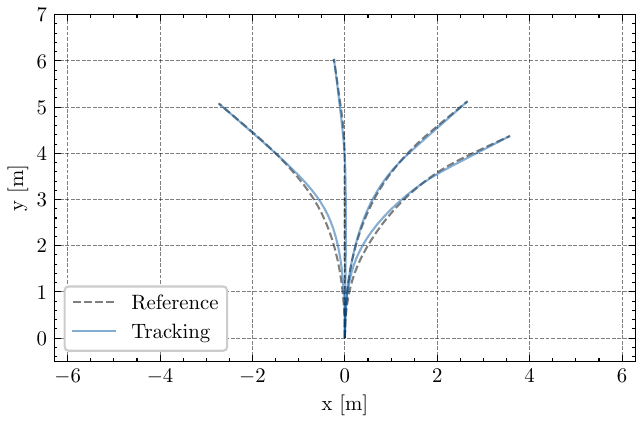}
  \vspace{-5mm}
  \caption{Driving controller performance evaluated in simulation.}
  \label{fig:saha_driving_result}
\end{figure}

The driving controller was designed using a backstepping approach and offers convergence guarantees subject to kinematic limits.
We thus evaluated its performance in simulation with the motion primitives used by the local planner as references.
Since the driving controller's tracking targets during deployment are always sampled from these primitives, this evaluation validates the driving controller's performance within the integrated system.
The tracking performance on selected motion primitives is shown in~\cref{fig:saha_driving_result} as an example.
The driving controller successfully tracks the reference trajectories with small errors.
The average cross-track error, measured as the distance between the vehicle and the reference trajectory, across all motion primitives is 3.86\,cm, which is 16\% lower than a simple pure pursuit controller~\cite{Hu24MotionPrimitives}.

\vspace{-5mm}
\subsubsection{Arm Controller}
\begin{figure}
  \centering
  \includegraphics[width=\linewidth]{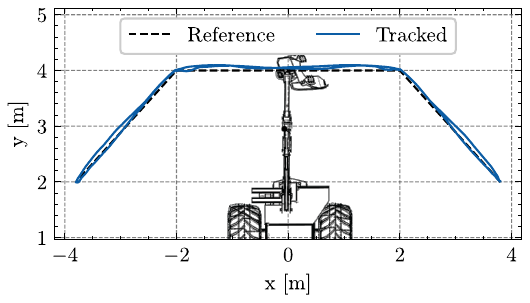}
  \vspace{-5mm}
  \caption{Arm controller performance evaluated on the real robot.}
  \label{fig:arm_waypoint_tracking}
\end{figure}

The adaptive arm controller was evaluated on the real robot by commanding the end-effector to follow a series of waypoints in 3D space.
The performance on a trajectory with four waypoints is shown in~\cref{fig:arm_waypoint_tracking}.
On average, the end-effector position tracking error is about 5\,cm up to end-effector speeds of 60\,cm/s~\cite{Nan24LearningAdaptive}.
As the gripper opening on the harvesting tool is about 40\,cm, this accuracy is sufficient for performing harvesting tasks on smaller-diameter trees.

\begin{figure}
  \centering
  \includegraphics[width=\linewidth]{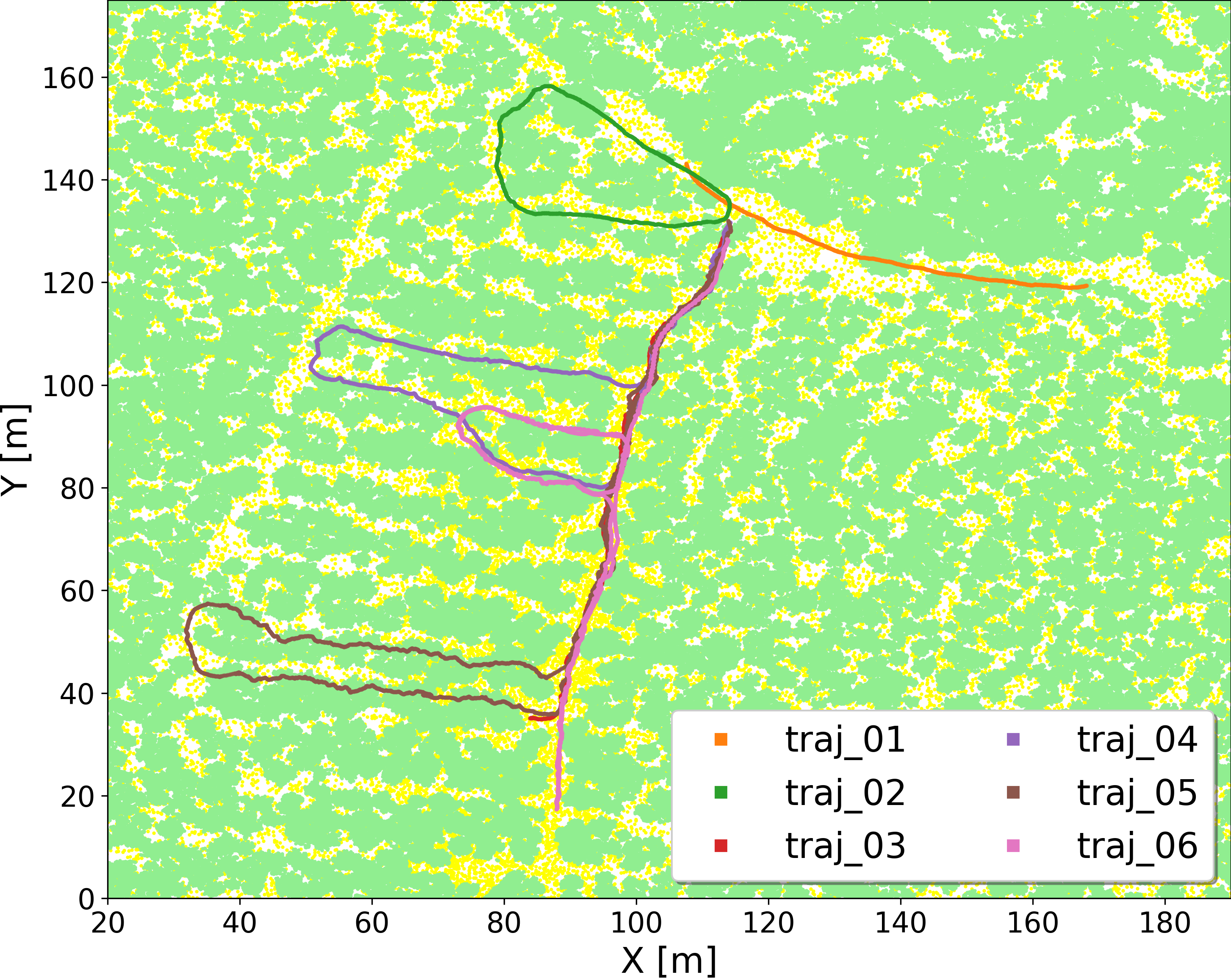}
  \vspace{-3mm}
  \caption{Visualization of SAHA semi-autonomous trajectories from field tests in Porvoo, Finland. The navigation is carried out autonomously, whereas an operator takes over for the tree cutting operations.}
  \label{fig:saha_porvoo_trajs}
\end{figure}

\subsection{Field Deployment}
We conducted a series of semi-autonomous and autonomous field deployments to test the proposed approach under real-world conditions.
\subsubsection{Teleoperation with Autonomous Driving}
The semi-autonomous deployment of SAHA was tested for extended periods of time in real thinning missions.
In these experiments, an operator selects target trees for thinning in a previously built forest inventory by providing a list of waypoints SAHA needs to visit.
SAHA localizes itself in the point-cloud map of the task area and drives autonomously towards the next waypoint.
Once a target tree is within reach of SAHA's arm, the robot stops and gives control to an operator who uses teleoperation to finish the cutting operation.
This represents a realistic use case for semi-autonomous thinning, where the operator can focus solely on the cutting operation while the robot navigates autonomously.
This also boosts productivity by allowing one operator to manage multiple harvesters simultaneously, with one being teleoperated during cutting and others driving autonomously to the next tree.

We carry out such semi-autonomous missions during autumn and winter near Porvoo, Finland.
These trials took place amid ongoing thinning operations, providing the most realistic test environment.
The site featured a mixture of open and dense stands, relatively flat terrain with gentle slopes and bumps, and pre-existing service trails left by other machinery. As visualized in~\cref{fig:saha_porvoo_trajs}, the test routes included straight public roads bordering the forest (\texttt{traj\_01}) as well as more complex trajectories within the forest that combined straight paths with turning maneuvers (\texttt{traj\_03}--\texttt{traj\_06}).
During these field tests, SAHA successfully drove autonomously for a total distance of 7.22\,km on forest roads and off-road terrain, consistently avoiding collisions. The longest continuous autonomous run without human intervention was 762\,m, and the average speed during autonomous navigation was 0.46\,m/s.
As a reference, a human operator achieves an average speed of 0.45\,m/s in forest trails and 0.85\,m/s on paved roads when operating SAHA using the remote controller.

\vspace{-5mm}
\subsubsection{Autonomous Tree Cutting}

\begin{figure}
  \centering
  \includegraphics[width=\linewidth]{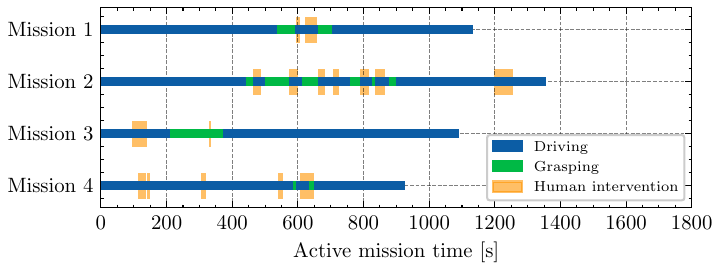}
  \vspace{-5mm}
  \caption{Summary of four cutting missions by SAHA in Evo, Finland. The missions involve navigation from the starting point to the target tree, reaching and grasping the tree with the harvesting head, followed by driving back to the base. The experiments' durations, phases, and human interventions are indicated in the figure. The interventions are primarily for briefly repositioning SAHA when it fails to grasp the tree, after which the operator manually drives SAHA back a few meters to enable a new autonomous approach. For clarity, we only count active operating time and exclude the idling periods when SAHA is autonomous but not in action, e.g., when the operator is selecting the target.}
  \label{fig:final_eval_quantative}
  \vspace{-5mm}
\end{figure}

\begin{figure*}
  \centering
  \includegraphics[width=\linewidth]{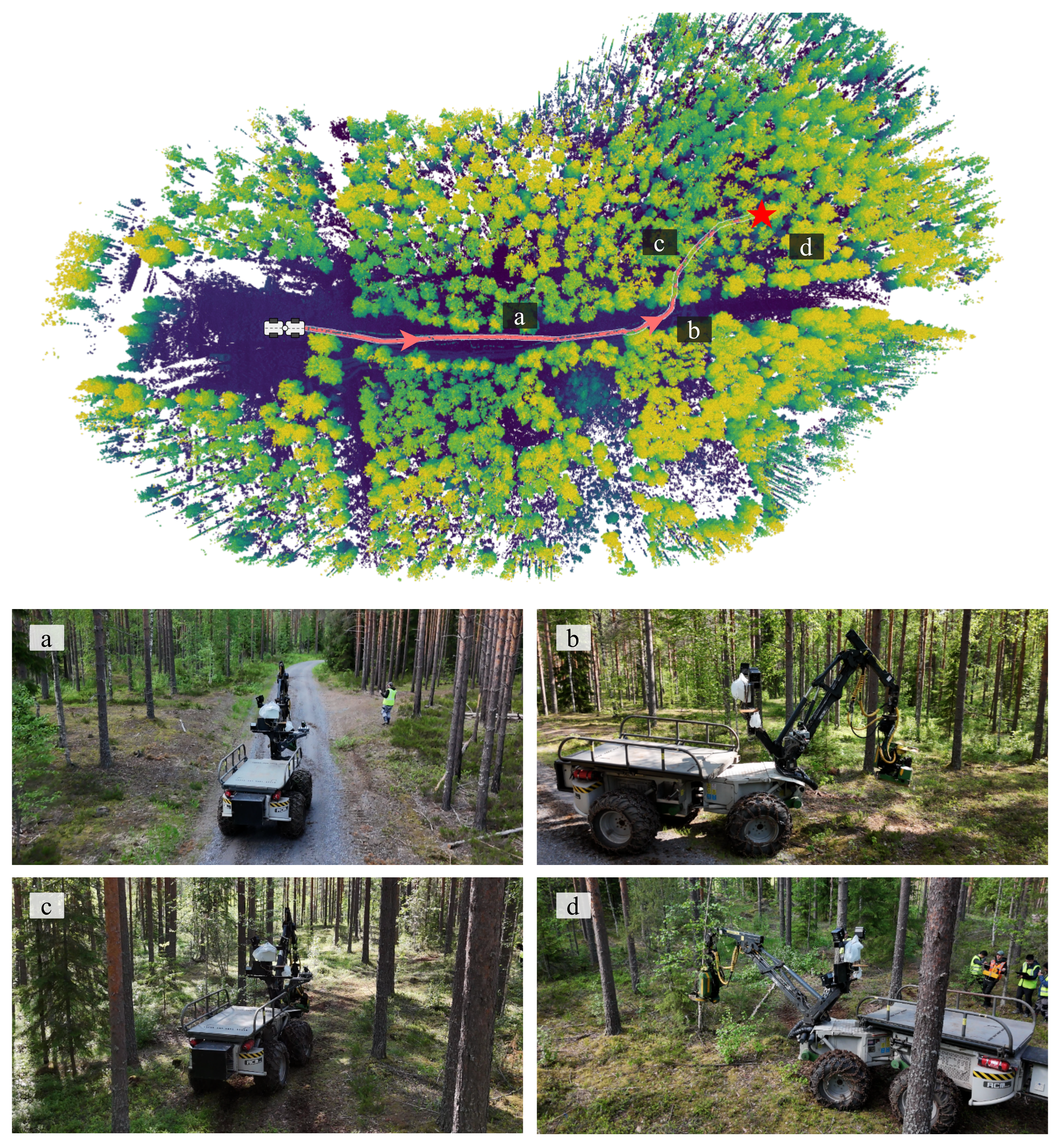}
  \vspace{-5mm}
  \caption{Visualization of a complete cutting mission by SAHA. The mission starts with SAHA on a forest road and an operator-selected target tree~(marked by the star). (a) SAHA autonomously drives along the road toward the target tree, (b) then leaves the road to enter the forest through a traversable gap. (c) SAHA navigates through the forest, avoiding trees and other obstacles. (d) Upon reaching the target tree, SAHA stops and grasps it with the harvesting head, ready for the cutting operation.}
  \label{fig:experiment_integrated}
\end{figure*}

The integrated autonomous cutting missions were conducted in a forest near Evo, Finland.
Over a two-day test period, SAHA completed 4 missions involving navigation and tree cutting.
We summarize the results from these experiments in~\cref{fig:final_eval_quantative}.
In each mission, SAHA's operating time was labeled by different operating stages and if human interventions were engaged.
In mission 1, 2, and 4, SAHA starts from more than 120\,m away from the target tree, thus requiring signigificantly longer time for navigation compared to mission 3, where it starts from about 40\,m away from the target.
In all the missions, SAHA finished the task of navigating to and grasping the target tree.
However, some human interventions were required during the process.
The most common intervention was to re-position SAHA when it fails to grasp the target tree on the first attempt, which happened in all four missions.
When this occurs, the operator manually drives SAHA back a few meters to let it re-attempt the final approach and grasp.
Such interventions were needed multiple times in mission 2 before SAHA successfully grasped the tree.
Overall, human interventions were required once every 281.6\,s on average during the missions, contributing to 9.23\% of the total operating time.

A panoramic view of mission 3 is visualized in~\cref{fig:experiment_integrated}.
The mission begins with SAHA positioned on a forest road, with the target tree about 150\,m away.
SAHA autonomously drives along the road towards the target tree.
When it detects a sufficiently wide traversable gap near the target, it leaves the road and enters the forest.
The robot then manages to navigate through the trees, avoiding collisions and reaching a waypoint just in front of the target tree.
Upon arrival, SAHA autonomously moves its arm to grasp the target tree with the harvesting head.
Once the human operator confirms that the tree is securely held, they command SAHA to perform the cut.

\section{Discussion}
Our experiments demonstrate the feasibility of the SAHA system for autonomously performing selective thinning tasks with a robotic harvester.
The system successfully integrates multiple technical components, including state estimation, perception, navigation, and control. These components combine to form a coherent autonomous forestry platform, which we validated through field trials in real forest environments.

Compared to previous autonomous forestry systems~\cite{Jelavic22AutonomousRobotic}, SAHA offers several key advantages that make it more suitable for practical deployment in selective thinning operations.
SAHA is built on a small-scale harvester platform, providing a more realistic and cost-effective solution for selective thinning.
While earlier work relied on expensive research platforms such as modified 12-ton hydraulic excavators, our lighter-weight harvester provides sufficient capability for first thinning operations while being substantially more accessible to forestry operators.
This smaller scale also reduces soil compaction and environmental impact, both critical considerations in sustainable forest management.

SAHA incorporates onboard traversability analysis and performs real-time obstacle avoidance instead of relying on offline planning using pre-surveyed data.
This enables SAHA to respond dynamically to unexpected obstacles, including humans, which significantly improves the operational safety of autonomous forest thinning.
By integrating learning-based traversability classification with probabilistic mapping, the system operates effectively in previously unseen forest areas without requiring detailed prior terrain knowledge.

SAHA has undergone a solid field validation, including the first demonstrated autonomous robotic thinning operation in a real forest environment. 
The conducted trials validate the system's performance under realistic operating conditions, including dense undergrowth, variable terrain, and challenging lighting conditions that are typical of forest thinning operations.

\subsection{Limitations and Future Directions}
Despite these successes, our field deployment has revealed several limitations that suggest important directions for future development.

The point cloud-based traversability analysis performs effectively in many scenarios but faces two key limitations. First, it struggles to distinguish geometrically ambiguous cases such as traversable tall grass versus non-traversable small trees.
Both appear as vertical structures with similar density and height in LiDAR data, creating ambiguity.
The method also cannot distinguish grassland from service trails due to the lack of geometric cues, even though service trails should be preferred to minimize soil disturbance.
These limitations also arise partly because the traversability classification network is trained on an external dataset collected from a different forest and sensor setup~\cite{malladi2025icra}. Moreover, this dataset was labeled for panoptic segmentation rather than explicit traversability. Integrating traversability-aware training data from the target forest would likely improve results considerably.
Future work should also explore combining LiDAR data with visual information to provide richer cues for semantic understanding, enabling better differentiation between ambiguous terrain types.

The current pure local planning approach, while robust for obstacle avoidance, is limiting in that it does not fully utilize available map information effectively.
For example, the planner can select suboptimal paths where it attempts to cut through forested areas although established service routes are accessible nearby.
Future iterations should incorporate global path preferences that prioritize established trails and service roads.
Additionally, the system could benefit from maintaining a memory of past trajectories and encouraging reuse of previously traversed paths, further reducing cumulative soil compaction.

\begin{figure}[t]
  \centering
  \includegraphics[width=\linewidth]{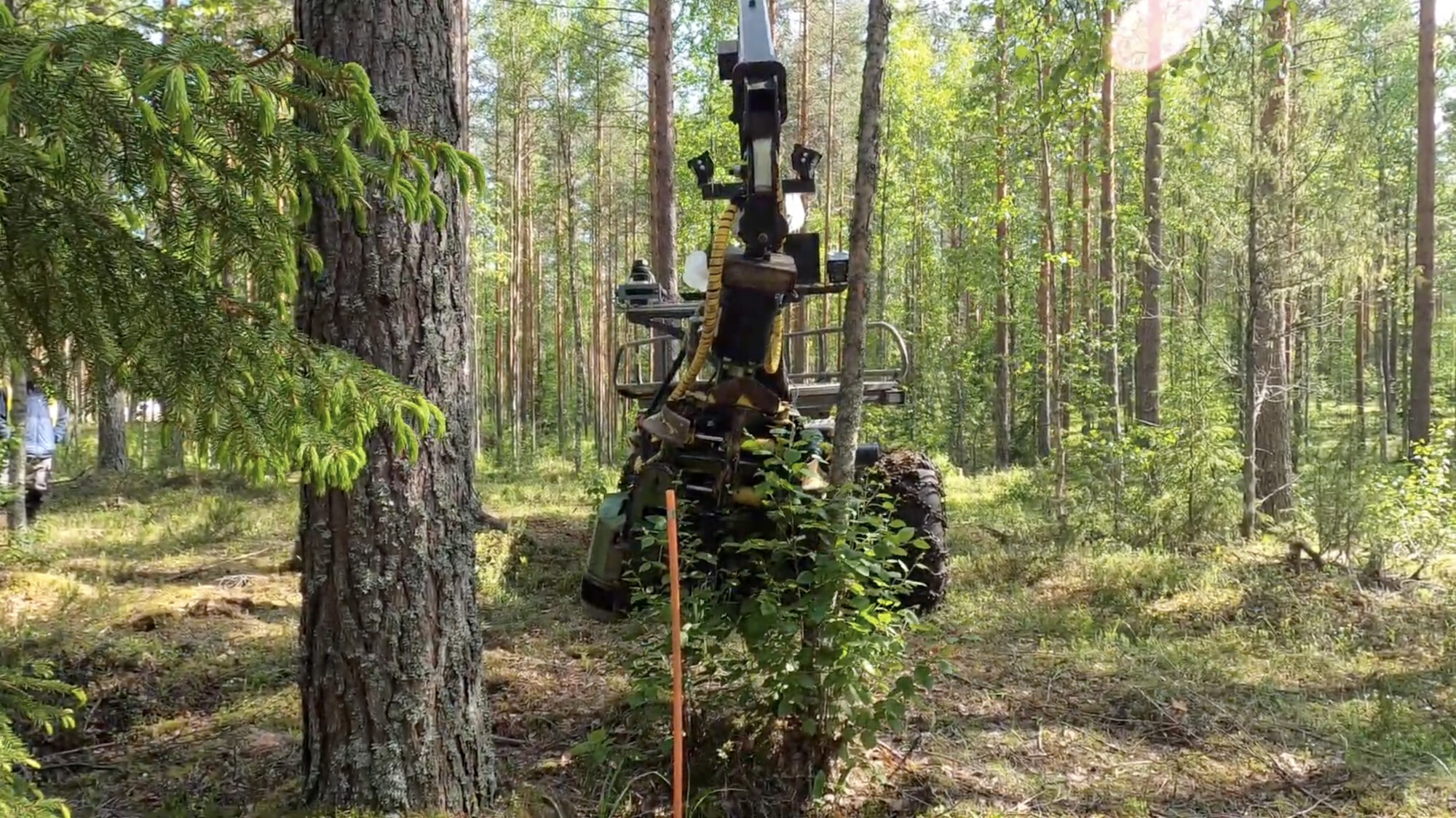}
  \caption{Example of a tree grasping failure caused by an offset between the controller's target tree location and the actual tree position. The onboard sensor's view of the target tree is obstructed by the boom and grapple.}
  \label{fig:tree_fov_blocked}
  \vspace{-4mm}
\end{figure}

Our approach of reaching target trees based solely on global map coordinates is not sufficiently robust and poses challenges for precision grasping operations.
While our odometry system exhibits relatively low drift, this drift can in the end still affect tree grasping performance.
Currently, SAHA is localized within the global reference map once at mission start. An improvement could involve completely switching to pose tracking in the prior map. However, this can increase memory and computational complexity, especially on embedded compute due to the larger map, and introduces further challenges in adapting to changes in the environment.
Alternatively, given the low drift in odometry, infrequent relocalization may also be sufficient.
An even more reliable approach for reaching and grasping target trees would involve detecting the trees using onboard sensors as the robot approaches, then performing the grasping based on this local perception.
Our current sensor setup cannot effectively support this capability because the boom and grapple obstruct the view of the tree, as shown in~\cref{fig:tree_fov_blocked}.
This would require additional perception capabilities, such as a secondary camera mounted on the gripper to provide close-range visual feedback during grasping maneuvers.

\section{Conclusion}
The presented SAHA system demonstrates the feasibility of autonomous forest thinning on a compact, field-deployable harvester platform.
Building on a 4.5-ton machine with targeted hardware modifications, the system integrates robust state estimation, learning-based traversability classification fused within a probabilistic mapping framework, a motion-primitive planner tailored to cluttered, under-canopy environments, and controllers for chassis balancing, vehicle path following, and hydraulic arm actuation.
Various field trials, including autonomous robotic thinning operations in a real forest environment, validate that SAHA can navigate dense stands, avoid obstacles, and position the harvesting head for tree cutting.

Beyond the integrated system performance, the individual components exhibit reliable operation under realistic conditions.
The odometry runs online on embedded compute resources, providing stable pose tracking even in canopy-occluded terrain.
The compact MinkUNet-based traversability estimator generalizes from an external dataset and, when accumulated probabilistically, yields actionable local maps that enable safe navigation.
The receding-horizon motion-primitive planner combined with the backstepping driving controller achieve accurate path following. Meanwhile, the adaptive hydraulic arm controller attains end-effector centimeter-level accuracy, enabling grasping of smaller-diameter trees.

The field deployments also reveal limitations that guide future work.
LiDAR-only traversability estimation can struggle with ambiguous scenarios and fails to distinguish service trails from similar-looking grasslands. Grasping based solely on global map coordinates can be offset by localization errors and scene changes.
Addressing these challenges will require multimodal perception that fuses LiDAR and vision to enable richer semantic understanding, incorporation of global path preferences and operational memory to favor established trails and previously traversed routes, and, finally, close-range sensing to re-detect target trees locally and adjust the grasp during the final approach.
Longer missions conducted across seasons and sites, improved human-in-the-loop supervision, and broader environmental impact evaluations are further steps toward practical deployment.

SAHA provides a practical blueprint for autonomous forest thinning. It combines a compact platform with integrated perception, planning, and control, capable of operating in unstructured forest environments with minimal prior knowledge.
By coupling local autonomy with targeted supervision and continuing to strengthen semantic perception, global navigation, and manipulation precision, autonomous harvesting can progress from controlled demonstrations toward reliable, scalable use in sustainable forest management.


\ifthenelse{\boolean{anonymous}}%
{} 
{ 
\section*{ACKNOWLEDGMENT}
The development of the SAHA robot is a large multidisciplinary effort and has received support from many people and organizations.
We extend our gratitude to Simon Kerscher, Arto Hongisto, Tun Kapgen, Aurel Kelterborn, Konrad Meyer, and Cedric Weibel for their support on the hardware development and integration.
We also acknowledge Tun Kapgen, Simon Kerscher, Jiangpeng Hu, Julian Nubert, Pol Eyschen, and Louis Wiesmann for contributions to the software stack.
Special thanks to Filippo Spinelli, Edo Jelavic, and Pascal Egli for sharing their insights in the development of the system.

The field experiments of the SAHA robot have been conducted as part of the integrated field tests under the Digiforest project (\url{https://digiforest.eu/}).
For the smooth execution of the field trials, we thank Stefan Leutenegger, Maurice Fallon, Henri Riihim\"aki, Jukka Maksimainen, and the staff at Prefor Oy for their professional organization.
The authors would like to thank all project partners for their invaluable support during the experiments.
In particular, we express our gratitude to Nived Chebrolu for providing the map of the test site.
}

\bibliographystyle{bibliography/IEEEtran}
\bibliography{bibliography/references_reformated}

@string{acc = {Proc.~of the IEEE American Control Conf.~(ACC)}}

@string{ar = {Autonomous Robots}}

@string{arxiv = {arXiv preprint}}

@string{compag = {Computers and Electronics in Agriculture}}

@string{corl = {Proc.~of the Conf.~on Robot Learning (CoRL)}}

@string{cvpr = {Proc.~of the IEEE/CVF Conf.~on Computer Vision and Pattern
                 Recognition (CVPR)}}

@string{cvprold = {Proc.~of the IEEE Conf.~on Computer Vision and Pattern
                    Recognition (CVPR)}}

@string{fps = {Frontiers in Plant Science}}

@string{icar = {Proc.~of the Intl.~Conf.~on Advanced Robotics (ICAR)}}

@string{icra = {Proc.~of the IEEE Intl.~Conf.~on Robotics \& Automation (ICRA)}}

@string{ijrr = {Intl.~Journal~of Robotics Research (IJRR)}}

@string{iros = {Proc.~of the IEEE/RSJ Intl.~Conf.~on Intelligent Robots and
                 Systems (IROS)}}

@string{jfr = {Journal of Field Robotics (JFR)}}

@string{jprs = {ISPRS Journal of Photogrammetry and Remote Sensing (JPRS)}}

@string{pers = {Photogrammetric Engineering and Remote Sensing (PE\&RS)}}

@string{ral = {IEEE Robotics and Automation Letters (RA-L)}}

@string{ram = {IEEE Robotics and Automation Magazine (RAM)}}

@string{rs = {Remote Sensing}}

@string{rss = {Proc.~of Robotics: Science and Systems (RSS)}}

@string{sensors = {Sensors}}

@string{siggraph = {Proc.~of the Intl.~Conf.~on Computer Graphics and
                     Interactive Techniques (SIGGRAPH)}}

@string{tpami = {IEEE Trans.~on Pattern Analysis and Machine Intelligence
                  (TPAMI)}}

@string{tra = {IEEE Trans.~on Robotics and Automation}}

@string{tro = {IEEE Trans.~on Robotics (TRO)}}

@string{tfr = {IEEE Trans.~on Field Robotics}}

@string{mee = {Methods in Ecology and Evolution}}

@string{rse = {Remote Sensing of Environment}}

@inproceedings{Casseau24MarkerlessAerialTerrestrial,
  author = {Casseau, Beno{\^i}t and Chebrolu, Nived and Mattamala, Matias and Frei{\ss}muth, Leonard and Fallon, Maurice},
  title = {Markerless {{Aerial-Terrestrial Co-Registration}} of {{Forest Point
           Clouds}} Using a {{Deformable Pose Graph}}},
  booktitle = iros,
  year = {2024},
}

@article{Cateanu24PerformanceEvaluation,
  title = {{Performance Evaluation of Real-Time Kinematic Global Navigation
           Satellite Systems under Forest Canopy}},
  author = {C{\u a}{\c t}eanu, M. and Marinescu, R. and R{\u a}dulescu, C. and
            Stoica, M.},
  year = {2024},
  journal = sensors,
  volume = {24},
  number = {13},
  pages = {1289},
}

@article{Chen23EndtoendAutonomous,
  author = {Chen, Li and Wu, Penghao and Chitta, Kashyap and Jaeger, Bernhard
            and Geiger, Andreas and Li, Hongyang},
  journal = tpami,
  number = {12},
  pages = {10164-10183},
  title = {{End-to-End Autonomous Driving: Challenges and Frontiers}},
  volume = {46},
  year = {2024},
}

@article{Corke01SteeringKinematics,
  title = {{Steering Kinematics for a Center-Articulated Mobile Robot}},
  author = {Corke, P.I. and Ridley, P.},
  year = {2001},
  journal = tra,
  volume = {17},
  number = {2},
  pages = {215--218},
}

@article{Deng18LearningbasedHierarchical,
  title = {{A Learning-Based Hierarchical Control Scheme for an Exoskeleton
           Robot in Human--Robot Cooperative Manipulation}},
  author = {Deng, Mingdi and Li, Zhijun and Kang, Yu and Chen, CL Philip and Chu
            , Xiaoli},
  year = {2018},
  journal = {IEEE Trans.~on Cybernetics},
  volume = {50},
  number = {1},
  pages = {112--125},
}

@article{DeSantis97ModelingPathtracking,
  title = {{Modeling and Path-Tracking for a Load-Haul-Dump Mining Vehicle}},
  author = {DeSantis, R. M.},
  year = {1997},
  journal = {Journal of Dynamic Systems, Measurement, and Control},
  volume = {119},
  number = {1},
  pages = {40--47},
}

@inproceedings{Dharmadhikari20MotionPrimitivesbased,
  title = {{Motion Primitives-Based Path Planning for Fast and Agile Exploration
           Using Aerial Robots}},
  booktitle = icra,
  author = {Dharmadhikari, Mihir and Dang, Tung and Solanka, Lukas and Loje,
            Johannes and Nguyen, Huan and Khedekar, Nikhil and Alexis, Kostas},
  year = {2020},
}

@article{Dubins57CurvesMinimal,
  title = {{On Curves of Minimal Length with a Constraint on Average Curvature,
           and with Prescribed Initial and Terminal Positions and Tangents}},
  author = {Dubins, Lester E},
  year = {1957},
  journal = {American Journal of Mathematics},
  volume = {79},
  number = {3},
  pages = {497--516},
}

@article{Egli22GeneralApproach,
  title = {A {{General Approach}} for the {{Automation}} of {{Hydraulic Excavator Arms Using Reinforcement Learning}}},
  author = {Egli, Pascal and Hutter, Marco},
  year = {2022},
  journal = ral,
  volume = {7},
  number = {2},
  pages = {5679--5686},
}

@article{Egli24ReinforcementLearningBased,
  title = {Reinforcement {{Learning-Based Bucket Filling}} for {{Autonomous
           Excavation}}},
  author = {Egli, Pascal and Terenzi, Lorenzo and Hutter, Marco},
  year = {2024},
  journal = tfr,
  volume = {1},
  pages = {170--191},
}

@inproceedings{Freissmuth24OnlineTree,
  title = {Online {{Tree Reconstruction}} and {{Forest Inventory}} on a {{Mobile Robotic System}}},
  booktitle = iros,
  author = {Frei{\ss}muth, Leonard and Mattamala, Matias and Chebrolu, Nived and
            Schaefer, Simon and Leutenegger, Stefan and Fallon, Maurice},
  year = {2024},
}

@inproceedings{Frey23FastTraversability,
  title = {Fast {{Traversability Estimation}} for {{Wild Visual Navigation}}},
  booktitle = rss,
  author = {Frey, Jonas and Mattamala, Matias and Chebrolu, Nived and Cadena,
            Cesar and Fallon, Maurice and Hutter, Marco},
  year = {2023},
}

@inproceedings{Gammell14InformedRRT,
  title = {Informed {{RRT}}: {{Optimal}} Sampling-Based Path Planning Focused
           via Direct Sampling of an Admissible Ellipsoidal Heuristic},
  booktitle = iros,
  author = {Gammell, Jonathan D and Srinivasa, Siddhartha S and Barfoot, Timothy
            D},
  year = {2014},
}

@inproceedings{Guo08OptimalTrajectory,
  title = {{Optimal Trajectory Generation for Nonholonomic Robots in Dynamic
           Environments}},
  booktitle = icra,
  author = {Guo, Yi and Tang, Tang},
  year = {2008},
}

@article{Ha02RoboticExcavation,
  title = {{Robotic Excavation in Construction Automation}},
  author = {Ha, Quang and Santos, M. and Nguyen, Quang and Rye, D. and {
            Durrant-Whyte}, H.},
  year = {2002},
  journal = ram,
  volume = {9},
  number = {1},
  pages = {20--28},
}

@article{Hart68FormalBasis,
  title = {{A Formal Basis for the Heuristic Determination of Minimum Cost Paths}},
  author = {Hart, Peter E and Nilsson, Nils J and Raphael, Bertram},
  year = {1968},
  journal = {IEEE Trans.~on Systems Science and Cybernetics},
  volume = {4},
  number = {2},
  pages = {100--107},
}

@inproceedings{Hu24MotionPrimitives,
  author = {Hu, Jiangpeng and Yang, Fan and Nan, Fang and Hutter, Marco},
  booktitle = iros,
  title = {{Motion Primitives Planning For Center-Articulated Vehicles}},
  year = {2024},
}

@inproceedings{Hutter15OptimalForce,
  title = {{Towards Optimal Force Distribution for Walking Excavators}},
  booktitle = icar,
  author = {Hutter, Marco and Leemann, Philipp and Stevsic, Stefan and Michel,
            Andreas and Jud, Dominic and Hoepflinger, Mark and Siegwart, Roland
            and Figi, Ruedi and Caduff, Christian and Loher, Markus and Tagmann,
            Stefan},
  year = {2015},
}

@article{Hutter17ForceControl,
  title = {Force {{Control}} for {{Active Chassis Balancing}}},
  author = {Hutter, Marco and Leemann, Philipp and Hottiger, Gabriel and Figi,
            Ruedi and Tagmann, Stefan and Rey, Gonzalo and Small, George},
  year = {2017},
  journal = {IEEE/ASME Trans.~on Mechatronics},
  volume = {22},
  number = {2},
  pages = {613--622},
}

@inproceedings{HwanJeon13OptimalMotion,
  title = {{Optimal Motion Planning with the Half-Car Dynamical Model for
           Autonomous High-Speed Driving}},
  booktitle = acc,
  author = {{hwan Jeon}, Jeong and Cowlagi, Raghvendra V. and Peters, Steven C.
            and Karaman, Sertac and Frazzoli, Emilio and Tsiotras, Panagiotis and
            Iagnemma, Karl},
  year = {2013},
}

@inproceedings{Jarin-Lipschitz21DispersionminimizingMotion,
  title = {{Dispersion-Minimizing Motion Primitives for Search-Based Motion
           Planning}},
  booktitle = icra,
  author = {{Jarin-Lipschitz}, Laura and Paulos, James and Bjorkman, Raymond and
            Kumar, Vijay},
  year = {2021},
}

@article{Jelavic22AutonomousRobotic,
  author = {Jelavic, Edo and Jud, Dominic and Egli, Pascal and Hutter, Marco},
  journal = {Field Robotics},
  title = {{Robotic Precision Harvesting: Mapping, Localization, Planning and
           Control for a Legged Tree Harvester}},
  volume = {2},
  year = {2022},
  pages = {1386--1431},
}

@inproceedings{Jelavic21CombinedSampling,
  title = {Combined {{Sampling}} and {{Optimization Based Planning}} for {{
           Legged-Wheeled Robots}}},
  booktitle = icra,
  author = {Jelavic, Edo and Farshidian, Farbod and Hutter, Marco},
  year = {2021},
}

@inproceedings{Jelavic22HarveriSmall,
  title = {Harveri: {{A Small}} ({{Semi-}}){{Autonomous Precision Tree
           Harvester}}},
  booktitle = {{ICRA Workshop on Innovation in Forestry Robotics: Research and Industry Adoption}},
  author = {Jelavic, Edo and Kapgen, Tun and Kerscher, Simon and Jud, Dominic
            and Hutter, Marco},
  year = {2022},
}

@article{Johns23FrameworkRobotic,
  title = {{A Framework for Robotic Excavation and Dry Stone Construction Using
           On-Site Materials}},
  author = {Johns, Ryan Luke and Wermelinger, Martin and Mascaro, Ruben and Jud,
            Dominic and Hurkxkens, Ilmar and Vasey, Lauren and Chli, Margarita
            and Gramazio, Fabio and Kohler, Matthias and Hutter, Marco},
  year = {2023},
  journal = {Science Robotics},
  volume = {8},
  number = {84},
  pages = {eabp9758},
}

@article{Jud17PlanningControl,
  title = {Planning and {{Control}} for {{Autonomous Excavation}}},
  author = {Jud, Dominic and Hottiger, Gabriel and Leemann, Philipp and Hutter,
            Marco},
  year = {2017},
  journal = ral,
  volume = {2},
  number = {4},
  pages = {2151--2158},
}

@article{Jud21HEAPAutonomous,
  title = {{{HEAP}} - {{The}} Autonomous Walking Excavator},
  author = {Jud, Dominic and Kerscher, Simon and Wermelinger, Martin and Jelavic
            , Edo and Egli, Pascal and Leemann, Philipp and Hottiger, Gabriel and
            Hutter, Marco},
  year = {2021},
  journal = {Automation in Construction},
  volume = {129},
  pages = {103783},
}

@article{LaHera24ExploringFeasibility,
  title = {{Exploring the Feasibility of Autonomous Forestry Operations: Results
           from the First Experimental Unmanned Machine}},
  author = {La Hera, Pedro and {Mendoza-Trejo}, Omar and Lindroos, Ola and
            Lideskog, H{\aa}kan and Lindb{\"a}ck, Torbj{\"o}rn and Latif, Saira
            and Li, Songyu and Karlberg, Magnus},
  year = {2024},
  journal = jfr,
  volume = {41},
  number = {4},
  pages = {942--965},
}

@article{Lee22PrecisionMotion,
  title = {Precision {{Motion Control}} of {{Robotized Industrial Hydraulic
           Excavators}} via {{Data-Driven Model Inversion}}},
  author = {Lee, Minhyeong and Choi, Hyelim and Kim, ChangU and Moon, Jihyun and
            Kim, Dongmok and Lee, Dongjun},
  year = {2022},
  journal = ral,
  volume = {7},
  number = {2},
  pages = {1912--1919},
}

@inproceedings{Low21PROMPTProbabilistic,
  title = {{{PROMPT}}: {{Probabilistic}} Motion Primitives Based Trajectory
           Planning},
  booktitle = rss,
  author = {L{\"o}w, Tobias and Bandyopadhyay, Tirthankar and Williams, Jason
            and Borges, Paulo VK},
  year = {2021},
}

@inproceedings{malladi2024icra,
  title = {Tree {{Instance Segmentation}} and {{Traits Estimation}} for {{
           Forestry Environments Exploiting LiDAR Data Collected}} by {{Mobile
           Robots}}},
  booktitle = icra,
  author = {Malladi, M.V.R. and Guadagnino, T. and Lobefaro, L. and Mattamala,
            M. and Griess, H. and Schweier, J. and Chebrolu, N. and Fallon, M.
            and Behley, J. and Stachniss, C.},
  year = {2024},
}

@article{Marchi18SustainableForest,
  title = {Sustainable {{Forest Operations}} ({{SFO}}): {{A}} New Paradigm in a
           Changing World and Climate},
  author = {Marchi, Enrico and Chung, Woodam and Visser, Rien and Abbas, Dalia
            and Nordfjell, Tomas and Mederski, Piotr S. and McEwan, Andrew and
            Brink, Michal and Laschi, Andrea},
  year = {2018},
  journal = {Science of The Total Environment},
  volume = {634},
  pages = {1385--1397},
}

@article{Mattamala25BuildingForest,
  title = {Building {{Forest Inventories With Autonomous Legged Robots}}---{{System}}, {{Lessons}}, and {{Challenges Ahead}}},
  author = {Mattamala, Mat{\'i}as and Chebrolu, Nived and Frey, Jonas and Frei{\ss}muth, Leonard and Oh, Haedam and Casseau, Benoit and Hutter, Marco and Fallon, Maurice},
  year = {2025},
  journal = tfr,
  volume = {2},
  pages = {418--436},
}

@article{Miki22LearningRobust,
  title = {{Learning Robust Perceptive Locomotion for Quadrupedal Robots in the
           Wild}},
  author = {Miki, Takahiro and Lee, Joonho and Hwangbo, Jemin and Wellhausen,
            Lorenz and Koltun, Vladlen and Hutter, Marco},
  year = {2022},
  journal = {Science Robotics},
  volume = {7},
  number = {62},
  pages = {eabk2822},
}

@article{Murtiyoso24VirtualForests,
  title = {{Virtual Forests: A Review on Emerging Questions in the Use and
           Application of {{3D}} Data in Forestry}},
  author = {Murtiyoso, Arnadi and Holm, Stefan and Riihim{\"a}ki, Henri and
            Krucher, Anna and Griess, Holger and Griess, Verena Christiane and
            Schweier, Janine},
  year = {2024},
  journal = {Intl. Journal of Forest Engineering},
  volume = {35},
  number = {1},
  pages = {29--42},
}

@article{Nan24LearningAdaptive,
  title = {Learning {{Adaptive Controller}} for {{Hydraulic Machinery Automation}}},
  author = {Nan, Fang and Hutter, Marco},
  year = {2024},
  journal = ral,
  volume = {9},
  number = {4},
  pages = {3972--3979},
}

@phdthesis{Nayl13ModelingControl,
  title = {{Modeling, Control and Path Planning for an Articulated Vehicle}},
  author = {Nayl, T.},
  year = {2013},
  school = {Lule{\aa} tekniska universitet},
}

@article{Pierzchala18MappingForests,
  title = {Mapping Forests Using an Unmanned Ground Vehicle with {{3D LiDAR}}
           and Graph-{{SLAM}}},
  author = {Pierzcha{\l}a, Marek and Gigu{\`e}re, Philippe and Astrup, Rasmus},
  year = {2018},
  journal = compag,
  volume = {145},
  pages = {217--225},
}

@inproceedings{Pivtoraiko11KinodynamicMotion,
  title = {{Kinodynamic Motion Planning with State Lattice Motion Primitives}},
  booktitle = iros,
  author = {Pivtoraiko, Mihail and Kelly, Alonzo},
  year = {2011},
}

@article{Reeds90OptimalPaths,
  title = {{Optimal Paths for a Car That Goes Both Forwards and Backwards}},
  author = {Reeds, James and Shepp, Lawrence},
  year = {1990},
  journal = {Pacific Journal of Mathematics},
  volume = {145},
  number = {2},
  pages = {367--393},
}

@article{Spinelli25LargeScale,
  title = {{Large Scale Robotic Material Handling: Learning, Planning, and Control}},
  author = {Spinelli, Filippo A. and Zhai, Yifan and Nan, Fang and Egli, Pascal
            and Nubert, Julian and Bleumer, Thilo and Miller, Lukas and Hofmann,
            Ferdinand and Hutter, Marco},
  volume = {arXiv:2508.09003},
  journal = arxiv,
  year = {2025},
}

@article{Terenzi24AutonomousExcavation,
  title = {Toward {{Autonomous Excavation Planning}}},
  author = {Terenzi, Lorenzo and Hutter, Marco},
  year = {2024},
  journal = tfr,
  volume = {1},
  pages = {292--317},
}

@article{Tremblay20AutomaticThreedimensional,
  title = {{Automatic Three-Dimensional Mapping for Tree Diameter Measurements
           in Inventory Operations}},
  author = {Tremblay, Jean-Fran{\c c}ois and B{\'e}land, Martin and Gagnon,
            Richard and Pomerleau, Fran{\c c}ois and Gigu{\`e}re, Philippe},
  year = {2020},
  journal = jfr,
  volume = {37},
  number = {8},
  pages = {1328--1346},
}

@misc{U.S.BureauOfLaborStatistics24CivilianOccupations,
  title = {{Civilian Occupations with High Fatal Work Injury Rates, 2023}},
  author = {{U.S. Bureau of Labor Statistics}},
  year = {2024},
  url = {
         https://www.bls.gov/charts/census-of-fatal-occupational-injuries/civilian-occupations-with-high-fatal-work-injury-rates.htm
         },
}

@inproceedings{Weigand21HybridDataDriven,
  title = {Hybrid {{Data-Driven Modelling}} for {{Inverse Control}} of {{
           Hydraulic Excavators}}},
  booktitle = iros,
  author = {Weigand, Jonas and Raible, Julian and Zantopp, Nico and Demir, Ozan
            and Trachte, Adrian and Wagner, Achim and Ruskowski, Martin},
  year = {2021},
}

@article{Wu25ApplicationSLAMbased,
  title = {Application of {{SLAM-based}} Mobile Laser Scanning in Forest
           Inventory: {{Methods}}, Progress, Challenges, and Perspectives},
  author = {Wu, Y. and Zhong, S. and Ma, Y. and Zhang, Y. and Liu, M.},
  year = {2025},
  journal = {Forests},
  volume = {16},
  number = {6},
  pages = {920},
}

@article{Zhang20FalcoFast,
  title = {Falco: {{Fast}} Likelihood-Based Collision Avoidance with Extension
           to Human-Guided Navigation},
  author = {Zhang, J. and Hu, C. and others},
  year = {2020},
  journal = jfr,
  volume = {37},
  number = {8},
  pages = {1300--1313},
}

@article{Zhang21AutonomousExcavator,
  title = {{An Autonomous Excavator System for Material Loading Tasks}},
  author = {Zhang, Liangjun and Zhao, Jinxin and Long, Pinxin and Wang, Liyang
            and Qian, Lingfeng and Lu, Feixiang and Song, Xibin and Manocha,
            Dinesh},
  year = {2021},
  journal = {Science Robotics},
  volume = {6},
  number = {55},
  pages = {eabc3164},
}

@inproceedings{choy2019cvpr,
  author = {C. Choy and J. Gwak and S. Savarese},
  title = {{4D Spatio-Temporal ConvNets: Minkowski Convolutional Neural Networks}},
  booktitle = cvpr,
  year = 2019,
}

@article{malladi2025arxiv,
  author = {Malladi, M.V.R. and Guadagnino, T. and Lobefaro, L. and Stachniss,
            C.},
  title = {{A Robust Approach for LiDAR-Inertial Odometry Without
           Sensor-Specific Modeling}},
  journal = arxiv,
  year = {2025},
  volume = {arXiv:2509.06593},
}

@inproceedings{museth2013siggraph,
  title = {{OpenVDB: An Open-source Data Structure and Toolkit for
           High-resolution Volumes}},
  author = {Museth, Ken and Lait, Jeff and Johanson, John and Budsberg, Jeff and
            Henderson, Ron and Alden, Mihai and Cucka, Peter and Hill, David and
            Pearce, Andrew},
  booktitle = {Proc.~of the ACM SIGGRAPH Courses},
  year = {2013},
}

@inproceedings{malladi2025icra,
  author = {Malladi, M.V.R. and Chebrolu, N. and Scacchetti, I. and Lobefaro, L. and Guadagnino, T. and Casseau, B. and Oh, H. and Frei{\ss}muth, L. and Karppinen, M. and Schweier, J. and Leutenegger, S. and Behley, J. and Stachniss, C. and Fallon, M.},
  title = {{DigiForests: A Longitudinal LiDAR Dataset for Forestry Robotics}},
  booktitle = icra,
  year = {2025},
}

@article{mersch2023ral,
  author = {Mersch, B. and Guadagnino, T. and Chen, X. and Vizzo, I. and Behley,
            J. and Stachniss, C.},
  title = {{Building Volumetric Beliefs for Dynamic Environments Exploiting
           Map-Based Moving Object Segmentation}},
  journal = ral,
  volume = {8},
  number = {8},
  pages = {5180--5187},
  year = {2023},
}

@article{wiesmann2024arxiv-ba,
  title = {{Efficient LiDAR Bundle Adjustment for Multi-Scan Alignment Utilizing
           Continuous-Time Trajectories}},
  author = {Wiesmann, Louis and Marks, Elias and Gupta, Saurabh and Guadagnino,
            Tiziano and Behley, Jens and Stachniss, Cyrill},
  journal = arxiv,
  volume = {arXiv:2412.11760},
  year = {2024},
}

@inproceedings{guadagnino2025icra,
  author = {Guadagnino, T. and Mersch, B. and Vizzo, I. and Gupta, S. and
            Malladi, M.V.R. and Lobefaro, L. and Doisy, G. and Stachniss, C.},
  title = {{Kinematic-ICP: Enhancing LiDAR Odometry with Kinematic Constraints
           for Wheeled Mobile Robots Moving on Planar Surfaces}},
  booktitle = icra,
  year = 2025,
}

@inproceedings{geiger2012cvpr,
  author = {A. Geiger and P. Lenz and R. Urtasun},
  title = {{Are we ready for Autonomous Driving? The KITTI Vision Benchmark
           Suite}},
  booktitle = cvprold,
  year = {2012},
}

@article{wulder2012rse,
  author = {Michael A. Wulder and Joanne C. White and Ross F. Nelson and Erik
            Næsset and Hans Ole Ørka and Nicholas C. Coops and Thomas Hilker and
            Christopher W. Bater and Terje Gobakken},
  journal = rse,
  number = {},
  pages = {196--209},
  title = {{Lidar sampling for large-area forest characterization: A review}},
  volume = {121},
  year = {2012},
}

@article{dalponte2016mee,
  author = {Michele Dalponte and David A. Coomes},
  journal = mee,
  number = {10},
  pages = {1236--1245},
  title = {Tree-centric mapping of forest carbon density from airborne laser
           scanning and hyperspectral data},
  volume = {7},
  year = {2016},
}

@article{sun2022fps,
  author = {Chenxin Sun and Chengwei Huang and Huaiqing Zhang and Bangqian Chen
            and Feng An and Liwen Wang and Ting Yun},
  journal = fps,
  title = {{Individual Tree Crown Segmentation and Crown Width Extraction From a
           Heightmap Derived From Aerial Laser Scanning Data Using a Deep
           Learning Framework}},
  volume = {13},
  pages={914974},
  year = {2022},
}

@article{roussel2020rse,
  author = {Jean-Romain Roussel and David Auty and Nicholas C. Coops and Piotr
            Tompalski and Tristan R.H. Goodbody and Andrew Sánchez Meador and
            Jean-François Bourdon and Florian {de Boissieu} and Alexis Achim},
  journal = rse,
  number = {},
  pages = {112061},
  title = {{lidR: An R package for analysis of Airborne Laser Scanning (ALS)
           data}},
  volume = {251},
  year = {2020},
}

@article{li2012pers,
  author = {Wenkai Li and Qinghua Guo and Marek K. Jakubowski and Maggi Kelly},
  journal = pers,
  number = {1},
  pages = {75--84},
  title = {{A New Method for Segmenting Individual Trees from the Lidar Point
           Cloud}},
  volume = {78},
  year = {2012},
}

@article{du2019rs,
  author = {Shenglan Du and Roderik Lindenbergh and Hugo Ledoux and Jantien
            Stoter and Liangliang Nan},
  journal = rs,
  number = {18},
  title = {{AdTree: Accurate, Detailed, and Automatic Modelling of Laser-Scanned
           Trees}},
  volume = {11},
  pages={2074},
  year = {2019},
}

@article{hannah2022essd,
  author = {Weiser, Hannah and Sch\"afer, Jannika and Winiwarter, Lukas and Kra\v{s}ovec, Nina and Fassnacht, Fabian E. and H\"ofle, Bernhard},
  journal = {Earth System Science Data},
  number = {7},
  pages = {2989--3012},
  title = {{Individual tree point clouds and tree measurements from
           multi-platform laser scanning in German forests}},
  volume = {14},
  year = {2022},
}

@article{burt2019mee,
  author = {Andrew Burt and Mathias Disney and Kim Calders},
  journal = mee,
  number = {3},
  pages = {438--445},
  title = {Extracting individual trees from lidar point clouds using treeseg},
  volume = {10},
  year = {2019},
}

@article{liang2018jprs,
  author = {Xinlian Liang and Juha Hyyppä and Harri Kaartinen and Matti
            Lehtomäki and Jiri Pyörälä and Norbert Pfeifer and Markus Holopainen
            and Gábor Brolly and Pirotti Francesco and Jan Hackenberg and Huabing
            Huang and Hyun-Woo Jo and Masato Katoh and Luxia Liu and Martin
            Mokroš and Jules Morel and Kenneth Olofsson and Jose Poveda-Lopez and
            Jan Trochta and Di Wang and Jinhu Wang and Zhouxi Xi and Bisheng Yang
            and Guang Zheng and Ville Kankare and Ville Luoma and Xiaowei Yu and
            Liang Chen and Mikko Vastaranta and Ninni Saarinen and Yunsheng Wang},
  journal = jprs,
  pages = {137--179},
  title = {International benchmarking of terrestrial laser scanning approaches
           for forest inventories},
  volume = {144},
  year = {2018},
}

@article{gonzalez2018mee,
  author = {Jose Gonzalez de Tanago and Alvaro Lau and Harm Bartholomeus and
            Martin Herold and Valerio Avitabile and Pasi Raumonen and Christopher
            Martius and Rosa C. Goodman and Mathias Disney and Solichin Manuri
            and Andrew Burt and Kim Calders},
  journal = mee,
  number = {2},
  pages = {223--234},
  title = {{Estimation of above-ground biomass of large tropical trees with
           terrestrial LiDAR}},
  volume = {9},
  year = {2018},
}

@article{liang2016jprs,
  author = {Xinlian Liang and Ville Kankare and Juha Hyyppä and Yunsheng Wang
            and Antero Kukko and Henrik Haggrén and Xiaowei Yu and Harri
            Kaartinen and Anttoni Jaakkola and Fengying Guan and Markus
            Holopainen and Mikko Vastaranta},
  journal = jprs,
  pages = {63--77},
  title = {Terrestrial laser scanning in forest inventories},
  volume = {115},
  year = {2016},
}

@article{donager2021rs,
  author = {Jonathon J. Donager and Andrew J. Sánchez Meador and Ryan C.
            Blackburn},
  journal = rs,
  number = {12},
  title = {{Adjudicating Perspectives on Forest Structure: How Do Airborne,
           Terrestrial, and Mobile Lidar-Derived Estimates Compare?}},
  volume = {13},
  year = {2021},
  pages={2297},
}

@article{puliti2023arxiv,
  author = {Stefano Puliti and Grant Pearse and Peter Surovy and Luke Wallace
            and Markus Hollaus and Maciej Wielgosz and Rasmus Astrup},
  title = {{FOR-instance: a UAV laser scanning benchmark dataset for semantic
           and instance segmentation of individual trees}},
  journal = arxiv,
  volume = {arXiv:2309.01279},
  year = 2023,
}

@article{schiefer2020jprs,
  author = {Felix Schiefer and Teja Kattenborn and Annett Frick and Julian Frey
            and Peter Schall and Barbara Koch and Sebastian Schmidtlein},
  journal = jprs,
  pages = {205--215},
  title = {{Mapping forest tree species in high resolution UAV-based RGB-imagery by means of convolutional neural networks}},
  volume = {170},
  year = {2020},
}

@inproceedings{fortin2022iros,
  author = {Jean-Michel Fortin and Olivier Gamache and Vincent Grondin and
            François Pomerleau and Philippe Giguère},
  booktitle = iros,
  title = {{Instance Segmentation for Autonomous Log Grasping in Forestry
           Operations}},
  year = {2022},
}

@article{vidanapathirana2024ijrr,
  author = {Vidanapathirana, Kavisha and Knights, Joshua and Hausler, Stephen
            and Cox, Mark and Ramezani, Milad and Jooste, Jason and Griffiths,
            Ethan and Mohamed, Shaheer and Sridharan, Sridha and Fookes, Clinton
            and Moghadam, Peyman},
  journal = ijrr,
  number = {4},
  pages = {532--549},
  title = {{WildScenes: A benchmark for 2D and 3D semantic segmentation in
           large-scale natural environments}},
  volume = {44},
  year = {2025},
}

@inproceedings{cordts2016cvpr,
  author = {Cordts, Marius and Omran, Mohamed and Ramos, Sebastian and Rehfeld, Timo and Enzweiler, Markus and Benenson, Rodrigo and Franke, Uwe and Roth, Stefan and Schiele, Bernt},
  title = {{The Cityscapes Dataset for Semantic Urban Scene Understanding}},
  booktitle = cvprold,
  year = {2016},
}

@inproceedings{behley2021icra,
  author = {J. Behley and A. Milioto and C. Stachniss},
  title = {{A Benchmark for LiDAR-Based Panoptic Segmentation Based on KITTI}},
  booktitle = icra,
  year = 2021,
}

@article{wisth2023tro,
  author = {David Wisth and Marco Camurri and Maurice Fallon},
  title = {{VILENS: Visual, Inertial, Lidar, and Leg Odometry for All-Terrain
           Legged Robots}},
  journal = tro,
  volume = 39,
  number = 1,
  year = 2023,
  pages = {309--326},
}

@article{krisanski2021rs,
  author = {Sean Krisanski and Mohammad Sadegh Taskhiri and Susana Gonzalez
            Aracil and David Herries and Paul Turner},
  journal = rs,
  number = {8},
  title = {{Sensor Agnostic Semantic Segmentation of Structurally Diverse and
           Complex Forest Point Clouds Using Deep Learning}},
  volume = {13},
  year = {2021},
  pages={1413},
}

@article{ramezani2022arxiv,
  author = {Ramezani, Milad and Khosoussi, Kasra and Catt, Gavin and Moghadam,
            Peyman and Williams, Jason and Borges, Paulo and Pauling, Fred and
            Kottege, Navinda},
  journal = arxiv,
  title = {{Wildcat: Online Continuous-Time 3D Lidar-Inertial SLAM}},
  volume = {arXiv:2205.12595},
  year = {2022},
}

@article{shen2025arxiv,
  author = {Shen, Yanqing and Tuna, Turcan and Hutter, Marco and Cadena, Cesar
            and Zheng, Nanning},
  journal = arxiv,
  title = {{ForestLPR: LiDAR Place Recognition in Forests Attentioning Multiple
           BEV Density Images}},
  volume = {arXiv:2503.04475},
  year = {2025},
}

@inproceedings{oh2024iros,
  author = {Haedam Oh and Nived Chebrolu and Matias Mattamala and Leonard Frei{\ss}muth and Maurice Fallon},
  title = {{Evaluation and Deployment of LiDAR-based Place Recognition in Dense
           Forests}},
  booktitle = iros,
  year = 2024,
}

@article{cheng2023arxiv,
  title = {{TreeScope: An Agricultural Robotics Dataset for LiDAR-Based Mapping
           of Trees in Forests and Orchards}},
  author = {Derek Cheng and Fernando Cladera Ojeda and Ankit Prabhu and Xu Liu
            and Alan Zhu and Patrick Corey Green and Reza Ehsani and Pratik
            Chaudhari and Vijay Kumar},
  year = 2023,
  journal = arxiv,
  volume = {arXiv:2310.02162},
}

@article{bai2022ral,
  author = {Bai, Chunge and Xiao, Tao and Chen, Yajie and Wang, Haoqian and
            Zhang, Fang and Gao, Xiang},
  journal = ral,
  number = {2},
  pages = {4861--4868},
  title = {{Faster-LIO: Lightweight Tightly Coupled Lidar-Inertial Odometry
           Using Parallel Sparse Incremental Voxels}},
  volume = {7},
  year = {2022},
}

@inproceedings{cao2022icra,
  author = {Cao, Chao and Zhu, Hongbiao and Yang, Fan and Xia, Yukun and Choset,
            Howie and Oh, Jean and Zhang, Ji},
  booktitle = icra,
  title = {{Autonomous Exploration Development Environment and the Planning
           Algorithms}},
  year = {2022},
}

@article{dixit2024fr,
  author = {Dixit, Anushri and Fan, David D. and Otsu, Kyohei and Dey, Sharmita
            and Agha-Mohammadi, Ali-Akbar and Burdick, Joel W.},
  journal = tfr,
  pages = {81--99},
  title = {{STEP: Stochastic Traversability Evaluation and Planning for
           Risk-Aware Navigation; Results From the DARPA Subterranean Challenge}},
  volume = {2},
  year = {2025},
}

@inproceedings{yang2021icra,
  author = {Yang, Bowen and Wellhausen, Lorenz and Miki, Takahiro and Liu, Ming
            and Hutter, Marco},
  booktitle = icra,
  title = {{Real-time Optimal Navigation Planning Using Learned Motion Costs}},
  year = {2021},
}

@inproceedings{bradley2015ricirs,
  author = {Bradley, David M. and Chang, Jonathan K. and Silver, David and
            Powers, Matthew and Herman, Herman and Rander, Peter and Stentz,
            Anthony},
  booktitle = iros,
  title = {Scene Understanding for a High-mobility Walking Robot},
  year = {2015},
}

@inproceedings{shaban2022rl,
  author = {Shaban, Amirreza and Meng, Xiangyun and Lee, JoonHo and Boots, Byron
            and Fox, Dieter},
  booktitle = corl,
  title = {{Semantic Terrain Classification for Off-Road Autonomous Driving}},
  year = {2022},
}

@article{mattamala2025ar,
  author = {Mattamala, Matias and Frey, Jonas and Libera, Piotr and Chebrolu,
            Nived and Martius, Georg and Cadena, Cesar and Hutter, Marco and
            Fallon, Maurice},
  journal = ar,
  number = {3},
  pages = {19},
  title = {Wild Visual Navigation: Fast Traversability Learning via Pre-Trained
           Models and Online Self-Supervision},
  volume = {49},
  year = {2025},
}

@article{hertzberg2013if,
  author = {Hertzberg, Christoph and Wagner, René and Frese, Udo and Schröder,
            Lutz},
  journal = {Information Fusion},
  number = {1},
  pages = {57--77},
  title = {Integrating generic sensor fusion algorithms with sound state
           representations through encapsulation of manifolds},
  volume = {14},
  year = {2013},
}

@article{feng2016rse,
  author = {Feng, Min and Sexton, Joseph O. and Huang, Chengquan and Anand,
            Anupam and Channan, Saurabh and Song, Xiao-Peng and Song, Dan-Xia and
            Kim, Do-Hyung and Noojipady, Praveen and Townshend, John R.},
  journal = rse,
  pages = {73--85},
  title = {{Earth science data records of global forest cover and change:
           Assessment of accuracy in 1990, 2000, and 2005 epochs}},
  volume = {184},
  year = {2016},
}

@article{dannunzio2015fem,
  author = {d’Annunzio, Rémi and Sandker, Marieke and Finegold, Yelena and Min,
            Zhang},
  journal = {Forest Ecology and Management},
  pages = {124--133},
  title = {{Projecting global forest area towards 2030}},
  volume = {352},
  year = {2015},
}

@article{giffen2025fem,
  author = {Giffen, R. Alec and Ryan, Colleen M. and Lowenstein, Frank and
            Perschel, Robert T. and Tyrrell, Mary L.},
  journal = {Forest Ecology and Management},
  pages = {122968},
  title = {{Redefining sustainable forestry for a climate emergency}},
  volume = {595},
  year = {2025},
}

@article{sporcic2025forests,
  author = {\v{S}por\v{c}i\'{c}, Mario and Landeki\'{c}, Matija and Pandur,
            Zdravko and Ba\v{c}i\'{c}, Marin and Mato\v{s}evi\'{c}, Matej and
            Mijo\v{c}, David and Mu\v{s}i\'{c}, Jusuf},
  journal = {Forests},
  number = {7},
  title = {Development and Evaluation of Strategic Directions for Strengthening
           Forestry Workforce Sustainability},
  volume = {16},
  year = {2025},
}

@article{dye2024cbm,
  author = {Dye, Alex W. and Houtman, Rachel M. and Gao, Peng and Anderegg,
            William R. L. and Fettig, Christopher J. and Hicke, Jeffrey A. and
            Kim, John B. and Still, Christopher J. and Young, Kevin and Riley,
            Karin L.},
  journal = {Carbon Balance and Management},
  number = {1},
  pages = {35},
  title = {Carbon, climate, and natural disturbance: a review of mechanisms,
           challenges, and tools for understanding forest carbon stability in an
           uncertain future},
  volume = {19},
  year = {2024},
}

@article{nan2025efficient,
  title={Efficient Model-Based Reinforcement Learning for Robot Control via Online Learning},
  author={Nan, Fang and Ma, Hao and Guan, Qinghua and Hughes, Josie and Muehlebach, Michael and Hutter, Marco},
  journal= arxiv,
  volume = {arXiv:2510.18518},
  year = {2025},
}

@article{guan2023ar,
  author = {Guan, Tianrui and He, Zhenpeng and Song, Ruitao and Zhang, Liangjun},
  journal = {Autonomous Robots},
  number = {6},
  pages = {695--714},
  title = {{TNES: terrain traversability mapping, navigation and excavation system for autonomous excavators on worksite}},
  volume = {47},
  year = {2023}
}

\ifthenelse{\boolean{anonymous}}%
{} 
{ 

} 

\vfill\pagebreak

\end{document}